%% file: main.tex
\documentclass[]{fairmeta}

\usepackage{tikz}
\usetikzlibrary{shadings}

\definecolor{sprcolor}{RGB}{28, 84, 166}
\definecolor{springgreen}{RGB}{54, 159, 91}
\definecolor{springpink}{RGB}{226, 93, 151}
\definecolor{springcyan}{RGB}{40, 150, 170}
\definecolor{springblue}{RGB}{34, 124, 194}
\definecolor{springgold}{RGB}{242, 176, 75}

\newcommand{\gaussianball}{\tikz[baseline=-0.62ex, x=1ex, y=1ex]{\fill[springblue, opacity=0.09, rotate around={18:(-0.72,-0.18)}] (-0.72,-0.18) ellipse (0.48 and 0.27); \fill[springgreen, opacity=0.10, rotate around={-22:(-0.42,-0.30)}] (-0.42,-0.30) ellipse (0.52 and 0.28); \fill[springpink, opacity=0.08, rotate around={48:(-0.30,0.04)}] (-0.30,0.04) ellipse (0.34 and 0.20); \draw[springblue!75!black, line width=0.04ex, opacity=0.58, ->] (-0.42,0.06) -- (-0.18,0.23); \draw[springgreen!75!black, line width=0.04ex, opacity=0.58, ->] (0.16,-0.12) -- (0.42,-0.28); \draw[springpink!75!black, line width=0.038ex, opacity=0.52, ->] (0.22,0.24) -- (0.10,0.52); \draw[springgold!75!black, line width=0.038ex, opacity=0.50, ->] (-0.06,0.36) -- (-0.30,0.48); \draw[springblue!60!springgreen!80!black, line width=0.038ex, opacity=0.50, ->] (0.54,0.02) -- (0.80,0.16); \fill[springblue, opacity=0.34, rotate around={18:(-0.42,0.06)}] (-0.42,0.06) ellipse (0.56 and 0.32); \fill[springgreen, opacity=0.34, rotate around={-22:(0.16,-0.12)}] (0.16,-0.12) ellipse (0.62 and 0.34); \fill[springpink, opacity=0.28, rotate around={48:(0.22,0.24)}] (0.22,0.24) ellipse (0.42 and 0.25); \fill[springgold, opacity=0.30, rotate around={-10:(-0.06,0.36)}] (-0.06,0.36) ellipse (0.36 and 0.20); \fill[springblue!70!springgreen, opacity=0.26, rotate around={28:(0.54,0.02)}] (0.54,0.02) ellipse (0.34 and 0.21); \draw[springblue!75!black, line width=0.045ex, opacity=0.70, rotate around={18:(-0.42,0.06)}] (-0.42,0.06) ellipse (0.56 and 0.32); \draw[springgreen!75!black, line width=0.045ex, opacity=0.70, rotate around={-22:(0.16,-0.12)}] (0.16,-0.12) ellipse (0.62 and 0.34); \draw[springpink!75!black, line width=0.045ex, opacity=0.60, rotate around={48:(0.22,0.24)}] (0.22,0.24) ellipse (0.42 and 0.25); \draw[springgold!75!black, line width=0.04ex, opacity=0.55, rotate around={-10:(-0.06,0.36)}] (-0.06,0.36) ellipse (0.36 and 0.20); \draw[springblue!70!springgreen!70!black, line width=0.04ex, opacity=0.55, rotate around={28:(0.54,0.02)}] (0.54,0.02) ellipse (0.34 and 0.21); \fill[white, opacity=0.45] (-0.48,0.17) circle (0.07); \fill[white, opacity=0.38] (0.04,0.03) circle (0.06); \fill[white, opacity=0.34] (0.24,0.30) circle (0.05);}}

\title{
\vspace{-1pt}
\gaussianball\hspace{0.25em}
\textbf{GaussianDream}: A Feed-Forward 3D Gaussian World Model for Robotic Manipulation
}

\author[2,3,1,*]{Zijian Zhang}
\author[2,3,1,*]{Yuqing Jiang}
\author[4]{Qian Cheng}
\author[5]{Xiaofan Li}
\author[6]{Si Liu}
\author[7]{Ding Zhao}
\author[8]{Ping Luo}
\author[4]{Weitao Zhou}
\author[8,1,\ddagger]{Haibao Yu}

\affiliation[1]{Tuojing Intelligence}
\affiliation[2]{University of Chinese Academy of Sciences}
\affiliation[3]{Institute of Automation, Chinese Academy of Sciences}
\affiliation[4]{Tsinghua University}
\affiliation[5]{Zhejiang University}
\affiliation[6]{Beihang University}
\affiliation[7]{Carnegie Mellon University}
\affiliation[8]{The University of Hong Kong}

\contribution[*]{equal contribution}
\contribution[\ddagger]{corresponding author}

\input{math_commands.tex}

\usepackage{graphicx}
\usepackage[table]{xcolor}
\usepackage{colortbl}

\usepackage{booktabs}
\usepackage{tabularx}
\usepackage{array}
\usepackage{multirow}
\usepackage{diagbox}
\usepackage{hhline}
\usepackage{longtable}
\usepackage{makecell}
\usepackage{siunitx}
\usepackage{adjustbox}
\usepackage{wrapfig}
\usepackage{caption}   
\newcolumntype{C}{>{\centering\arraybackslash}X}

\usepackage{amsmath,amsfonts,amssymb}
\usepackage{bm}
\usepackage{nicefrac}

\usepackage{enumitem}
\setlist[itemize]{leftmargin=*}

\usepackage{caption}
\crefname{figure}{Fig.}{Figs.}
\crefname{table}{Tab.}{Tabs.}

\usepackage{xspace}
\usepackage{calc}
\usepackage{etoolbox}
\usepackage{pifont}
\usepackage{fancyvrb}

\usepackage{titletoc}

\titlecontents{section}
[1.5em] %
{\addvspace{-0.5pt}} %
{\bfseries\contentslabel{2.3em}} %
{\hspace*{-2.3em}\bfseries} %
{\bfseries\titlerule*[.5pc]{.}\contentspage} %
\titlecontents{subsection}
[3.8em] %
{\addvspace{-2.2pt}} %
{\contentslabel{2.3em}}
{\hspace*{-2.3em}}
{\titlerule*[.5pc]{.}\contentspage}

\abstract{
Vision-language-action (VLA) policies have advanced language-conditioned robotic manipulation by transferring semantic priors from pretrained vision-language models to action generation. 
However, standard action-imitation learning often lacks sufficient modeling of explicit 3D spatial information, dense geometric supervision, and future environment evolution, all critical for precise robotic interaction. 
To address this, we propose \textbf{GaussianDream}, a feed-forward 3D Gaussian world-model plug-in. 
Specifically, we introduce learnable GaussianDream Queries in the encoder, enabling the model to capture current-frame 3D spatial structure and short-horizon future evolution. 
During training, the latent GaussianDream prefix is processed by a static reconstruction head and a future prediction head to produce current 3D Gaussian scene states and future Gaussian evolution states. 
The current branch is supervised by RGB rendering and depth, while the future branch uses future RGB, depth, and pseudo 3D scene-flow signals. 
During inference, GaussianDream discards all auxiliary heads and retains only the learned prefix to condition action generation, without test-time Gaussian reconstruction or future prediction. 
Experimental results demonstrate that GaussianDream achieves state-of-the-art performance across multiple robotic manipulation benchmarks, reaching \textbf{98.4\%} on LIBERO, \textbf{54.8\%} on RoboCasa Human-50, and \textbf{50.0\%} on real-robot tasks. 
Compared with existing 3D-enhanced VLA methods, GaussianDream achieves strong accuracy while providing higher inference efficiency than video-based world-model approaches.
}
\metadata[Code]{\url{https://github.com/TuojingAI/GaussianDream}}

\definecolor{lightgray}{rgb}{0.95, 0.95, 0.95}

\definecolor{baselinecolor}{gray}{.9}

\begin{document}
\maketitle 
\input{sec/1_intro}
\input{sec/2_related_work}
\input{sec/3_method}
\input{sec/4._experiments}
\input{sec/5._conclusion}

\clearpage
\bibliography{main}
\bibliographystyle{bibstyle}

\newpage
\setcounter{section}{0}
\input{sec/6._appendix}

\end{document}

%% file: math_commands.tex
\usepackage{amsmath,amsfonts,bm}
\usepackage{xcolor}

\def\eqref#1{equation~\ref{#1}}

\def\1{\bm{1}}

\DeclareMathAlphabet{\mathsfit}{\encodingdefault}{\sfdefault}{m}{sl}
\SetMathAlphabet{\mathsfit}{bold}{\encodingdefault}{\sfdefault}{bx}{n}

%% file: sec/1_intro.tex
\section{Introduction}
\label{sec:intro}

General-purpose robotic manipulation, wherein an agent physically interacts with its environment to realize language-guided goals, is a fundamental task in embodied intelligence.
Recently, Vision-Language-Action (VLA) models, e.g., RT-2~\citep{brohan2023rt2}, OpenVLA~\citep{kim2024openvla}, $\pi_0$~\citep{black2024pi0}, and $\pi_{0.5}$~\citep{physicalintelligence2025pi05}, have driven remarkable progress in this domain. 
These architectures leverage pre-trained Vision-Language Models (VLMs) to ground semantic priors directly into physical control loops, enhancing manipulation performance and task generalization~\citep{brohan2023rt2, octo2024, kim2024openvla, black2024pi0, physicalintelligence2025pi05}.

Despite their strong instruction-following capabilities and semantic generalization, standard VLA paradigms still face critical limitations in precise physical interaction, partly because their training is largely dominated by behavior cloning~\citep{black2024pi0, physicalintelligence2025pi05, qian2025geopredict, song2026vga}. 
A fundamental deficiency lies in the \textbf{spatial and geometric underspecification} of existing architectures. 
Since pre-trained VLMs predominantly operate on 2D pixel grids, 3D spatial structures and contact constraints are often encoded only \textit{implicitly} within visual latents and action labels, making the control loop vulnerable to subtle geometric execution errors such as shifted grasp points~\citep{sun2025geovla, li2025qdepthvla, li2025spatialforcing, qian2025geopredict, song2026vga}. 
Furthermore, this paradigm suffers from the \textbf{underutilization of dense visual and spatial supervision} embedded within image observations. 
While robot trajectories record rich, high-dimensional evidence of object layouts, appearance cues, and depth structures, standard action-imitation objectives mainly supervise the immediate control command at each time step, leaving many dense pixel-level geometric signals underexploited~\citep{ye2026dreamzero, kim2026cosmospolicy, li2026lingbotva, bi2025motus}. 
Consequently, existing VLA policies typically acquire environment dynamics only implicitly from action labels, rather than through explicit internal environment emulation and future-state supervision. 
Although emerging robotic world models~\citep{ye2026dreamzero, kim2026cosmospolicy, li2026lingbotva, bi2025motus} have validated that predicting future environment states can substantially benefit action learning and policy robustness, standard VLAs often lack an explicit mechanism to anticipate how the environment will evolve after interaction. 
This limits their capability to generalize across varying and complex manipulation scenes where execution outcomes depend on short-horizon state changes.

To address these limitations, recent advanced efforts generally fall into two categories, yet both remain sub-optimal for high-performance, real-time closed-loop control. 
On one hand, \textbf{3D-enhanced policy networks} attempt to explicitly inject geometric structures into VLA frameworks; however, they predominantly rely on \textit{statically anchoring} the current scene configuration via representations like point clouds or depth maps, while providing limited mechanisms for dynamic future environment emulation~\citep{sun2025geovla, deng2025stereovla, zhou2025vla4d, ni2025swiftvla, li2025qdepthvla, li2025spatialforcing, abouzeid2025geoawarevla, song2026vga}. 
On the other hand, emerging \textbf{robotic world models} introduce proactive environment reasoning by forecasting future states in pixel, latent, or action spaces~\citep{ye2026dreamzero, kim2026cosmospolicy, li2026lingbotva, bi2025motus, cen2025worldvla, yuan2026fast}. 
While a few recent exploratory works attempt to incorporate 3D Gaussian representations into predictive frameworks~\citep{kerbl20233d, lu2024manigaussian, lu2025gwm, qian2025geopredict}, seamlessly integrating these world models with mainstream VLA architectures poses severe structural and operational challenges. 
Because their generative predictions typically require iterative voxel optimization or heavy visual autoregressive rollouts, they introduce substantial computational overhead during inference, making them difficult to deploy in high-frequency robotic control loops~\citep{ye2026dreamzero, kim2026cosmospolicy, yuan2026fast}. 
Consequently, how to equip an actionable VLA model with unified capabilities of explicit 3D space grounding, dense supervision, and future anticipation, while maintaining an elegant and lightweight inference pipeline, remains a critically unresolved challenge.

To close these gaps, we present \textbf{GaussianDream}, a unified, feed-forward 3D Gaussian world model framework tailored for language-conditioned robotic manipulation.
Instead of operating in unconstrained hidden spaces or relying on heavy video rollouts, GaussianDream bridges the semantic strength of pre-trained VLAs with the geometric precision of 3D representations by reconstructing and predicting physical environment states within a structured 3D Gaussian space~\citep{kerbl20233d}.
Concretely, we design an asymmetric training-and-inference architecture that functions as a geometry-aware plug-in.
During training, a lightweight spatio-temporal reasoning encoder interacts with learnable queries to extract 3D-aware features from temporal observations~\citep{wang2025vggt}.
These representations are processed by two decoupled auxiliary components: a static Gaussian head for current scene reconstruction and a dynamic, horizon-conditioned Gaussian head for future geometric change prediction.
The pipeline is optimized end-to-end with dense pixel-level RGB rendering, depth estimation, and pseudo 3D scene-flow supervision, transforming ordinary robot trajectories into rich structured learning signals.
Crucially, during online inference, all auxiliary decoding heads are discarded.
The policy retains only a compact, information-rich GaussianDream prefix to condition action generation, bypassing test-time Gaussian decoding, geometric rendering, video rollout, and additional planning.
Extensive evaluations across LIBERO~\citep{liu2023libero}, RoboCasa Human-50~\citep{nasiriany2024robocasa}, and physical real-robot tasks demonstrate that GaussianDream achieves strong and highly competitive performance, including the best results on several spatially demanding settings, while showing tight spatio-temporal alignment between predictive future emulation and executable control commands.

In summary, the primary contributions of this work are:
\begin{itemize}[leftmargin=*,itemsep=1pt,topsep=2pt]
    \item \textbf{Unified 3D Gaussian World Model.}
    We introduce \mbox{GaussianDream}, a feed-forward framework that integrates language-conditioned VLA policies with structured 3D Gaussian representations for robotic manipulation.

    \item \textbf{Spatio-Temporal Representation and Dense Supervision.}
    GaussianDream addresses three key VLA bottlenecks in a unified design: explicit 3D spatial grounding, dense pixel-level supervision from robot trajectories, and short-horizon predictive environment emulation.

    \item \textbf{Efficient Asymmetric Plug-in.}
    GaussianDream uses full 3D reconstruction and future prediction for training-time supervision, while discarding auxiliary decoding heads at deployment to avoid test-time Gaussian decoding, rendering, video rollout, or an additional planner.

    \item \textbf{Empirical Validation.}
    Experiments on LIBERO~\citep{liu2023libero}, RoboCasa Human-50~\citep{nasiriany2024robocasa}, and real robots show strong and highly competitive performance, with analyses validating spatio-temporal alignment between predictive emulation and action generation.
\end{itemize}

\begin{figure*}[t]
    \centering
    \includegraphics[width=\textwidth]{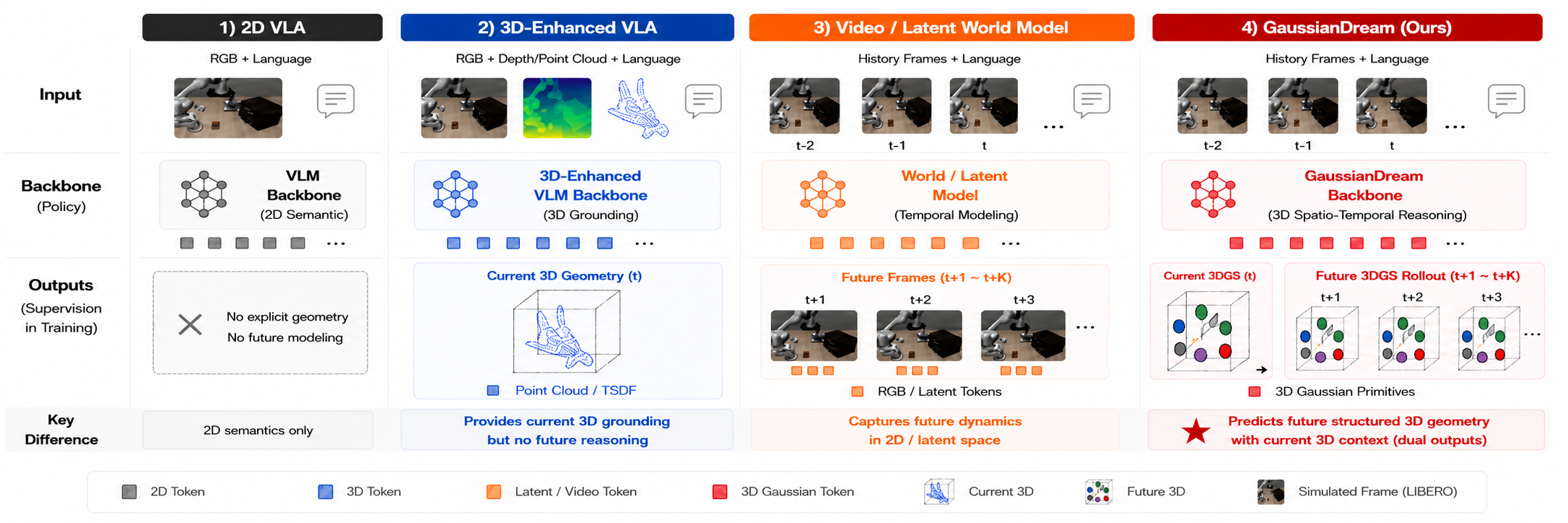}
    \caption{
    Comparison of manipulation policy paradigms.
    Compared with 2D policies, 3D-enhanced policies, and video/latent world models, GaussianDream uses current Gaussian reconstruction and future Gaussian prediction to learn structured 3D supervision from robot videos while retaining efficient prefix-based action generation.
    }
    \label{fig:intro_overview}
\end{figure*}

%% file: sec/2_related_work.tex
\section{Related Work}
\label{sec:related-work}

\paragraph{Vision-language robot policies.}
Vision-language-action (VLA) models adapt pretrained vision-language representations to robot control, enabling language-conditioned manipulation across diverse tasks.
RT-1~\citep{brohan2022rt1} and RT-2~\citep{brohan2023rt2} demonstrate the benefit of scaling robot data and transferring visual-language priors to action generation.
Octo~\citep{octo2024}, OpenVLA~\citep{kim2024openvla}, $\pi_0$~\citep{black2024pi0}, and $\pi_{0.5}$~\citep{physicalintelligence2025pi05} further advance open-vocabulary instruction following, continuous action modeling, and generalist policy learning.
Recent works improve action tokenization, deployability, speech conditioning, force awareness, and mobile manipulation~\citep{pertsch2025fast, wen2024tinyvla, shukor2025smolvla, zhao2025vlas, yu2026forcevla, wu2025momanipvla}.
While these models provide strong semantic priors, their supervision is still largely dominated by action imitation, where dense spatial-temporal evidence in robot videos is only indirectly encoded through visual latents and action labels.
GaussianDream preserves the VLA action interface while introducing explicit 3D Gaussian reconstruction, dense geometric supervision, and future Gaussian prediction.

\paragraph{3D-enhanced manipulation policies.}
A growing line of work improves robot policies by injecting explicit geometric information.
GeoVLA~\citep{sun2025geovla}, StereoVLA~\citep{deng2025stereovla}, VLA-4D~\citep{zhou2025vla4d}, SwiftVLA~\citep{ni2025swiftvla}, BridgeVLA~\citep{li2026bridgevla}, and Any3D-VLA~\citep{fan2026any3d} incorporate depth, stereo cues, 4D features, point clouds, or projected 3D representations for spatial grounding.
QDepth-VLA~\citep{li2025qdepthvla}, Spatial Forcing~\citep{li2025spatialforcing}, GeoAware-VLA~\citep{abouzeid2025geoawarevla}, and VGA~\citep{song2026vga} further explore depth supervision, geometric regularization, and geometry-centric backbones.
These methods show that explicit geometry benefits localization, spatial reasoning, and physically grounded control.
However, most use geometry to anchor the current scene, leaving post-interaction state evolution less explicit.
GeoPredict~\citep{qian2025geopredict} moves toward predictive geometry through future keypoint trajectories and depth-supervised Gaussian prediction.
GaussianDream instead unifies current reconstruction and horizon-conditioned future prediction as a training-time plug-in, turning robot trajectories into dense RGB, depth, and pseudo 3D scene-flow supervision.

\paragraph{World models for robotic manipulation.}
World models and World Action Models introduce future prediction as a way to learn richer temporal structure than one-step behavior cloning.
DreamZero~\citep{ye2026dreamzero}, Cosmos Policy~\citep{kim2026cosmospolicy}, LingBot-VA~\citep{li2026lingbotva}, Motus~\citep{bi2025motus}, WorldVLA~\citep{cen2025worldvla}, and Fast-WAM~\citep{yuan2026fast} predict future observations, actions, values, or latent states to improve long-horizon control and generalization.
Many of these methods model futures in RGB pixels, video latents, or action space, where the predicted state is not explicitly organized as actionable 3D geometry; autoregressive or diffusion-based rollouts may also increase inference cost.
Gaussian representations have recently been explored for manipulation or predictive geometry, including ManiGaussian~\citep{lu2024manigaussian}, GWM~\citep{lu2025gwm}, and GeoPredict~\citep{qian2025geopredict}.
In contrast, GaussianDream learns future anticipation directly in structured 3D Gaussian space during training, and discards auxiliary decoding heads at inference, retaining only a compact prefix for action generation.

%% file: sec/3_method.tex
\section{Method}
\label{sec:method}


\subsection{Overview of GaussianDream}

\begin{figure*}[t]
    \centering
    \includegraphics[width=\textwidth]{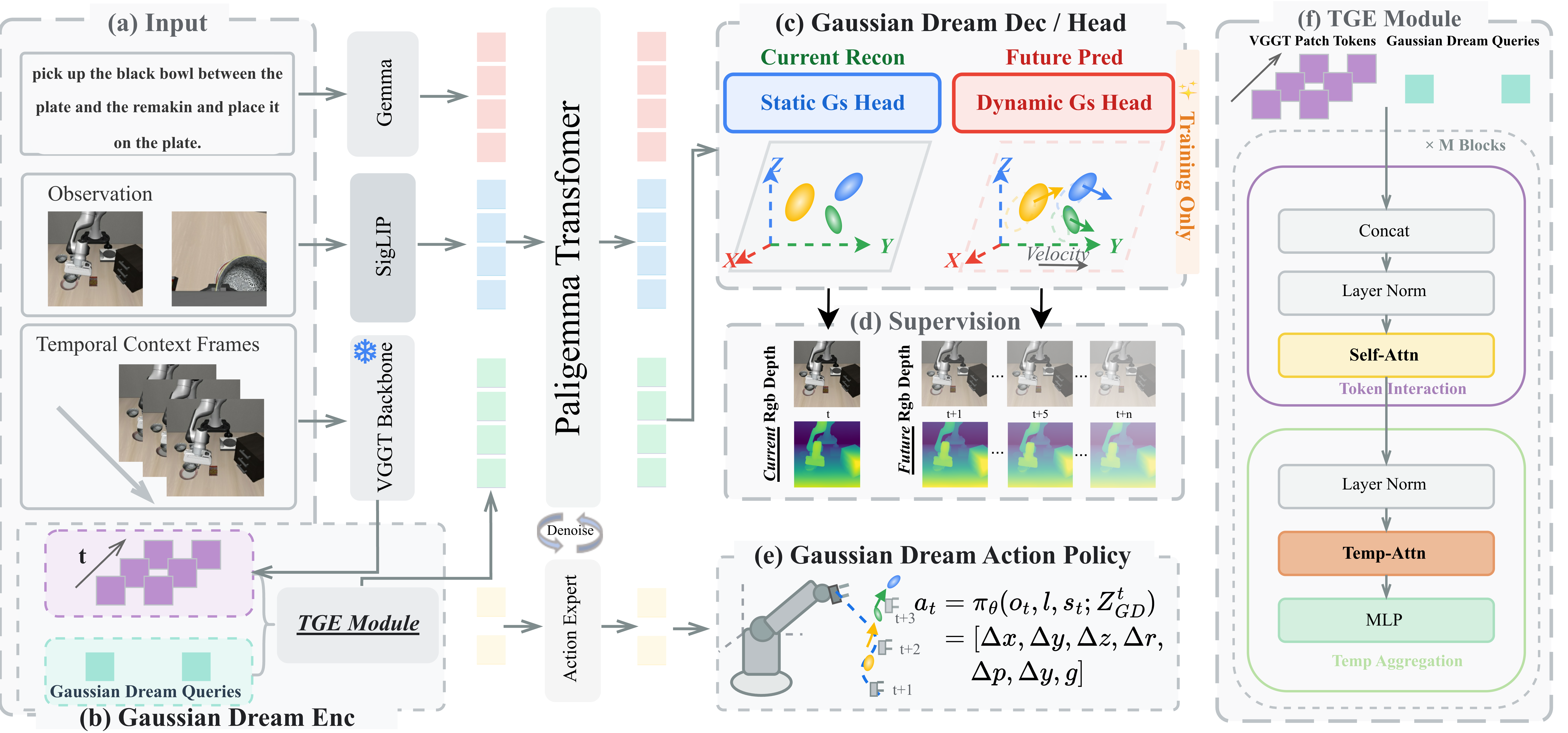}
    \caption{
    Overview of \textbf{GaussianDream}. 
    During training, temporal visual features and learnable GaussianDream queries produce a compact prefix, which is decoded into current and future 3D Gaussian states for RGB, depth, and pseudo 3D scene-flow supervision. 
    During inference, all auxiliary Gaussian decoding heads are discarded, and only the prefix conditions action generation.
    }
    \label{fig:method_overview}
\end{figure*}

Let $\mathbf{o}_t$ denote the current multi-view observation, $\mathbf{o}_{t-K:t}$ a short temporal history, $\mathbf{l}$ the language instruction, and $\mathbf{s}_t$ the robot state.
GaussianDream preserves the action interface of the base continuous-action policy, but augments its multimodal context with learned prefix tokens $\mathbf{Z}_t^{\mathrm{GD}}$.
In practice, the GaussianDream prefix is constructed from the temporal agent-view observation sequence, which provides consistent spatial context for 3D reconstruction and future prediction.
The prefix is generated by a temporal 3D-aware encoder:
\begin{equation}
    \mathbf{Z}_t^{\mathrm{GD}}
    =
    \mathcal{F}_{\omega}
    \left(
    \mathbf{o}_{t-K:t}, \mathbf{Q}_{\mathrm{GD}}
    \right),
\end{equation}

where $\mathbf{Q}_{\mathrm{GD}}$ are learnable GaussianDream queries and $\mathcal{F}_{\omega}$ is implemented by the Temporal Gaussian Evolution (TGE) module in Sec.~\ref{sec:method_future_prediction}.
Image, language, and GaussianDream tokens share the 2048-dimensional PaliGemma/Gemma-2B prefix space, while the action expert uses a 1024-dimensional hidden space.
We use three context frames, $\{t-10,t-5,t\}$, to provide sparse motion cues.

During training, the same prefix is decoded by two auxiliary Gaussian heads:
\begin{equation}
    \mathcal{G}_t
    =
    \mathcal{R}_{\phi}(\mathbf{Z}_t^{\mathrm{GD}}, \mathbf{o}_t),
    \qquad
    \hat{\mathcal{G}}_{t+\Delta}
    =
    \mathcal{D}_{\psi}(\mathcal{G}_t, \mathbf{Z}_t^{\mathrm{GD}}, \Delta).
\end{equation}
Here $\mathcal{R}_{\phi}$ maps the prefix to a renderable current Gaussian state $\mathcal{G}_t$, and $\mathcal{D}_{\psi}$ predicts horizon-conditioned future Gaussian changes.
These decoders serve only as training-time supervisory interfaces.
The policy conditions on the original multimodal inputs and the GaussianDream prefix:
\begin{equation}
    \mathbf{a}_t
    =
    \pi_{\theta}
    \left(
    \mathbf{o}_{t},\mathbf{l},\mathbf{s}_t;\mathbf{Z}_t^{\mathrm{GD}}
    \right).
\end{equation}
Thus, GaussianDream transfers reconstruction and prediction supervision into an inference-time prefix, rather than replacing the policy with a test-time simulator.

\subsection{Current Gaussian Reconstruction}
\label{sec:method_reconstruction}

The current reconstruction branch provides explicit 3D spatial grounding by requiring the GaussianDream prefix to represent a renderable 3D scene state.
The learned prefix $\mathbf{Z}_t^{\mathrm{GD}}$ serves as a compact 3D Gaussian latent representation that captures both current scene geometry and short-horizon interaction dynamics.
Its temporal evolution capability will be further modeled in Sec.~\ref{sec:method_future_prediction}.

We reshape the 1024 GaussianDream tokens into a spatially organized $32\times32$ latent grid, allowing the prefix to preserve spatial locality and dense geometric structure.
A Gaussian decoder then upsamples this latent grid into a dense feature map:
\begin{equation}
    \mathbf{F}_t^{\mathrm{G}}
    =
    \mathcal{B}_{\mathrm{G}}
    \left(
    \mathrm{Grid}(\mathbf{Z}_t^{\mathrm{GD}})
    \right),
    \qquad
    \mathbf{F}_t^{\mathrm{G}}\in\mathbb{R}^{256\times256\times128}.
\end{equation}
Here, the $256\times256$ spatial resolution corresponds to dense Gaussian layout prediction, while the 128-dimensional feature channels encode shared geometric and appearance information for subsequent Gaussian attribute estimation.

Lightweight prediction heads are then applied to construct the current Gaussian scene state.
The geometry head predicts depth, rotation, scale, and opacity, while the appearance head conditions on the current RGB observation $\mathbf{o}_t$ to estimate degree-1 spherical harmonics coefficients for Gaussian appearance:
\begin{equation}
    \mathbf{D}_t,\Theta_t^{\mathrm{geo}}
    =
    \mathcal{H}_{\mathrm{geo}}(\mathbf{F}_t^{\mathrm{G}}),
    \qquad
    \Theta_t^{\mathrm{app}}
    =
    \mathcal{H}_{\mathrm{app}}(\mathbf{F}_t^{\mathrm{G}},\mathbf{o}_t).
\end{equation}

We unproject $\mathbf{D}_t$ into Gaussian centers $\boldsymbol{\mu}_i^t$ and predict the corresponding Gaussian attributes $\theta_i^t$:
\begin{equation}
    \mathcal{G}_t
    =
    \mathcal{A}(\mathbf{D}_t,\Theta_t)
    =
    \{(\boldsymbol{\mu}_i^t,\theta_i^t)\}_{i=1}^{N_t},
\end{equation}
where $\Theta_t=(\Theta_t^{\mathrm{geo}},\Theta_t^{\mathrm{app}})$ and $N_t=256\times256$.

The reconstructed Gaussian scene is rendered for RGB and depth supervision.
By constraining the prefix to reconstruct explicit 3D geometry, this branch injects dense spatial structure into the policy representation and provides the static Gaussian template for future evolution prediction.

\subsection{Future Gaussian Prediction}
\label{sec:method_future_prediction}

Current reconstruction specifies the scene state at time $t$, but does not supervise how the scene evolves after interaction.
To expose dense temporal structure in robot videos, GaussianDream predicts short-horizon Gaussian evolution in structured 3D space.

Given $\mathbf{o}_{t-K:t}$, we extract multi-scale 3D-aware patch features using VGGT~\citep{wang2025vggt}.
Since VGGT learns strong static 3D priors through global frame attention but provides relatively weak supervision for temporal interaction dynamics, we introduce learnable GaussianDream queries to aggregate spatial geometry while modeling future evolution across time.
For each context frame, VGGT features are pooled into a $32\times32$ grid and projected into temporal tokens:
\begin{equation}
    \mathbf{P}_{t-K:t}^{(m)}
    =
    W_m\,
    \mathcal{P}_{32\times32}
    \left(
    \mathcal{E}_{\mathrm{VGGT}}^{(m)}(\mathbf{o}_{t-K:t})
    \right),
\end{equation}
where $m$ indexes the feature scale.

The Temporal Gaussian Evolution (TGE) module lets temporal tokens interact with learnable GaussianDream queries.
The 2048-dimensional queries are first projected into a 512-dimensional temporal space, fused with VGGT features from multiple frames, and finally projected back to the prefix space.
TGE contains 12 attention blocks with 8 heads, alternating frame-wise spatial interaction and time-slot temporal attention to jointly capture current 3D structure and future geometric evolution.
The current-step output is used as the GaussianDream prefix:
\begin{equation}
    \mathbf{Z}_t^{\mathrm{GD}}
    =
    \mathrm{Proj}_{512\rightarrow2048}
    \left[
    \mathrm{TGE}
    \left(
    \mathrm{Proj}_{2048\rightarrow512}(\mathbf{Q}_{\mathrm{GD}}),
    \{\mathbf{P}_{t-K:t}^{(m)}\}_{m=1}^{M}
    \right)_{t}
    \right].
\end{equation}

Given $\mathcal{G}_t=\{(\boldsymbol{\mu}_i^t,\theta_i^t)\}_{i=1}^{N_t}$, the dynamic head predicts horizon-conditioned center displacements:
\begin{equation}
\begin{aligned}
    \boldsymbol{\nu}_t^{(\Delta)}
    &=
    \mathcal{H}_{\mathrm{vel}}
    \left(
    \mathcal{B}_{\mathrm{pred}}(\mathbf{Z}_t^{\mathrm{GD}}),
    \mathbf{e}_{\Delta}
    \right),
    \qquad
    \Delta\mathbf{X}_t^{(\Delta)}
    =
    \alpha_{\Delta}\boldsymbol{\nu}_t^{(\Delta)}, \\
    \hat{\boldsymbol{\mu}}_i^{\,t+\Delta}
    &=
    \boldsymbol{\mu}_i^t+\Delta\mathbf{x}_i^{(\Delta)},
    \qquad
    \hat{\mathcal{G}}_{t+\Delta}
    =
    \left\{
    \left(
    \hat{\boldsymbol{\mu}}_i^{\,t+\Delta},
    \theta_i^t
    \right)
    \right\}_{i=1}^{N_t}.
\end{aligned}
\end{equation}

Here $\mathbf{e}_{\Delta}$ is a learnable horizon embedding and $\alpha_{\Delta}$ is a temporal scale factor.
Since robot interactions exhibit non-uniform motion patterns and time-dependent uncertainty, the horizon embedding $\mathbf{e}_{\Delta}$ enables the model to distinguish different dynamic modes across prediction horizons.
The future state reuses non-positional attributes $\theta_i^t$, focusing prediction on interaction-induced geometric changes while preserving the current Gaussian template.
We supervise horizons from $t+1$ to $t+5$.

\subsection{GaussianDream Training and Efficient Inference}

GaussianDream follows an asymmetric strategy: dense Gaussian reconstruction and prediction supervise training, while only the compact prefix is retained for online control.

\paragraph{Stage I: GaussianDream pretraining.}
We first train the reconstruction and prediction heads without action learning.
For each demonstration sequence, RGB frames are paired with pseudo depth and pseudo 3D scene-flow targets constructed from adjacent frames.
The GaussianDream objective combines current reconstruction and future prediction:
\begin{equation}
    \mathcal{L}_{\mathrm{GD}}
    =
    \underbrace{
    \lambda_{\mathrm{cur}}^{\mathrm{depth}}
    \mathcal{L}_{\mathrm{cur}}^{\mathrm{depth}}
    +
    \lambda_{\mathrm{cur}}^{\mathrm{render}}
    \mathcal{L}_{\mathrm{cur}}^{\mathrm{render}}
    }_{\mathcal{L}_{\mathrm{cur}}}
    +
    \underbrace{
    \sum_{\Delta\in\mathcal{H}}w_{\Delta}
    \left(
    \lambda_{\mathrm{depth}}
    \mathcal{L}_{\mathrm{depth}}^{(\Delta)}
    +
    \lambda_{\mathrm{render}}
    \mathcal{L}_{\mathrm{render}}^{(\Delta)}
    +
    \lambda_{\mathrm{flow}}
    \mathcal{L}_{\mathrm{flow}}^{(\Delta)}
    \right)
    }_{\mathcal{L}_{\mathrm{fut}}}.
\end{equation}
Depth and rendering losses supervise current and future Gaussian renderings, while the flow loss matches predicted center displacements to pseudo 3D scene flow under a validity mask.
This stage converts robot videos into dense spatial-temporal supervision rather than using them only as action-labeled trajectories.
The prediction horizon is gradually expanded for stable optimization.

\paragraph{Stage II: GaussianDream-conditioned policy learning.}
After pretraining, we jointly train the policy with auxiliary Gaussian losses to preserve the prefix's spatial-temporal structure.
With $\mathbf{c}_t=(\mathbf{o}_{t},\mathbf{l},\mathbf{s}_t;\mathbf{Z}_t^{\mathrm{GD}})$, the action loss is:
\begin{equation}
    \mathcal{L}_{\mathrm{act}}
    =
    \mathbb{E}_{\tau,\boldsymbol{\epsilon},\mathbf{a}_t^{\ast}}
    \left[
    \left\|
    \mathbf{v}_{\theta}
    \left(
    \tau\boldsymbol{\epsilon}
    +(1-\tau)\mathbf{a}_t^{\ast},
    \mathbf{c}_t,
    \tau
    \right)
    -
    (\boldsymbol{\epsilon}-\mathbf{a}_t^{\ast})
    \right\|_2^2
    \right],
\end{equation}
where $\boldsymbol{\epsilon}\sim\mathcal{N}(0,I)$ and $\tau$ is sampled from the flow-matching time distribution.
We optimize the joint objective $\mathcal{L}=\mathcal{L}_{\mathrm{act}}+\lambda_{\mathrm{GD}}\mathcal{L}_{\mathrm{GD}}$.
The action loss adapts the prefix to executable control, while the auxiliary Gaussian losses preserve its spatial-temporal structure.

\paragraph{Inference.}
At test time, GaussianDream encodes the temporal buffer into prefix tokens and injects them into the policy context.
All rendering, depth, velocity, and Gaussian decoding heads are discarded.
The base policy samples actions with its standard denoising procedure, without future rollout, test-time rendering, Gaussian decoding, or a separate planner.
Thus, GaussianDream retains training-time 3D supervision while preserving lightweight online control.

%% file: sec/4._experiments.tex
\section{Experiments}
\label{sec:experi}

We evaluate \textbf{GaussianDream} in simulation and real-world manipulation settings.
The experiments examine four questions: whether GaussianDream improves spatially precise manipulation over strong VLA, 3D-enhanced, and world-model baselines; whether the learned prefix transfers to physical execution; whether current reconstruction and future prediction are complementary; and whether rendering and depth losses provide useful dense supervision.

\subsection{Experimental Setup}

\paragraph{Simulation benchmarks.}
We evaluate GaussianDream on LIBERO~\citep{liu2023libero} and RoboCasa~\citep{nasiriany2024robocasa}.
For LIBERO, we follow the Spatial, Object, Goal, and Long protocols with 50 demonstrations and 50 evaluation trials.
For RoboCasa, we use the Human-50 few-shot setting over 24 long-horizon kitchen tasks, with 50 trials per task across five scenes.
Together, these benchmarks cover spatial reasoning, object grounding, goal-conditioned manipulation, and long-horizon household tasks.

\begin{figure*}[t]
    \centering
    \begin{minipage}[t]{0.48\textwidth}
        \centering
        \includegraphics[width=\linewidth]{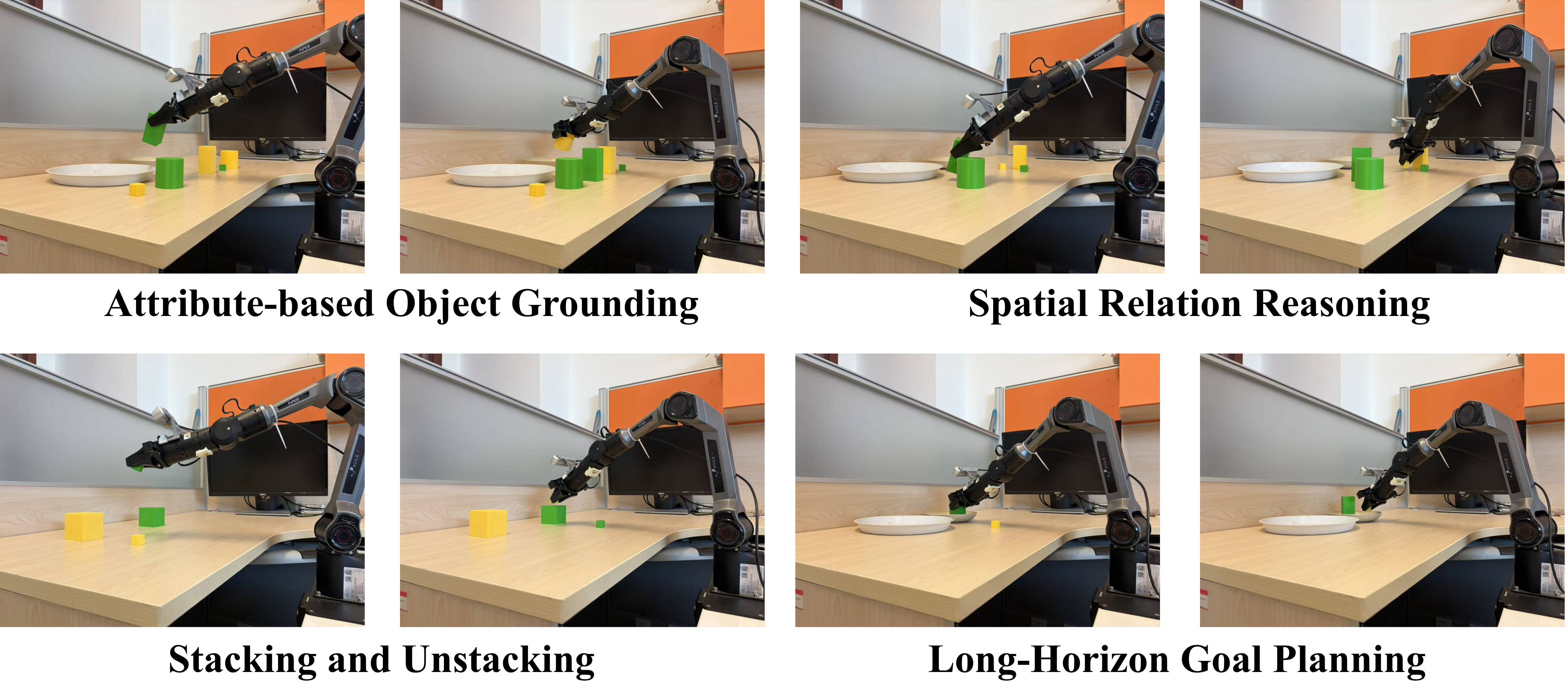}
        \caption{
        Real-robot tasks covering attribute grounding, spatial relations, stacking/unstacking, and long-horizon execution.
        }
        \label{fig:real_robot_tasks}
    \end{minipage}
    \hfill
    \begin{minipage}[t]{0.48\textwidth}
        \centering
        \includegraphics[width=\linewidth]{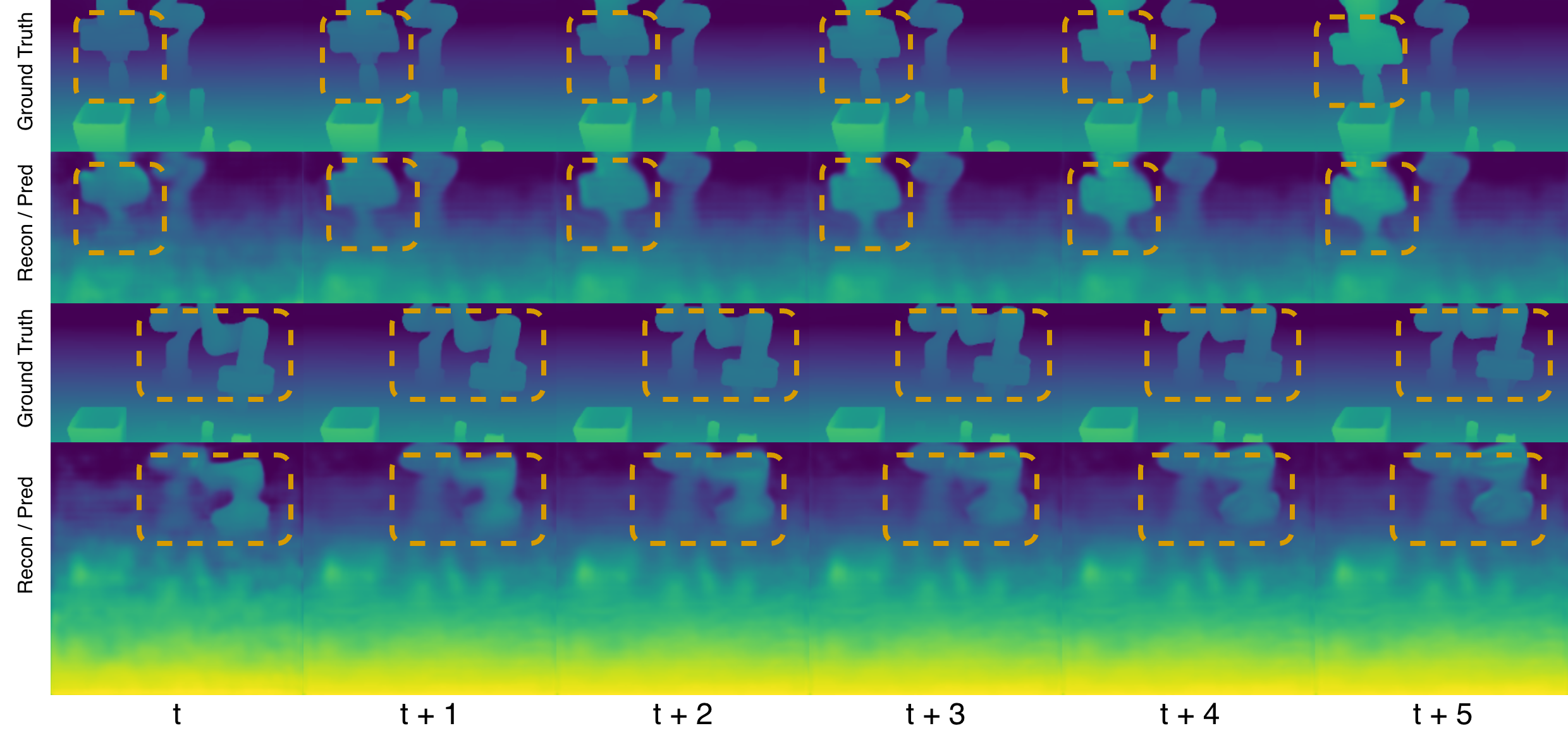}
        \caption{
        Ground-truth and GaussianDream depth-rendering visualizations analysis
        }
        \label{fig:quantitative_analysis_vis}
    \end{minipage}
\end{figure*}

\paragraph{Real-robot evaluation.}
We compare GaussianDream with the base $\pi_{0.5}$ policy~\citep{physicalintelligence2025pi05} on a physical robot.
As shown in Fig.~\ref{fig:real_robot_tasks}, the tasks cover attribute grounding, spatial relations, stacking/unstacking, and long-horizon execution.
These scenarios evaluate whether the GaussianDream prefix remains effective under camera noise, embodiment mismatch, and physical execution errors.

\subsection{Main Results}

\paragraph{LIBERO.}
GaussianDream achieves strong performance on LIBERO~\citep{liu2023libero}, especially on spatial grounding and goal-conditioned execution.
As shown in Tab.~\ref{tab:libero_main_results}, it reaches a 98.4\% average success rate and obtains the best Spatial and Goal scores.
For implementation-consistent comparison, we report the PyTorch implementation of Spatial Forcing~\citep{li2025spatialforcing}.
GaussianDream outperforms several 3D-enhanced policies, including GeoVLA~\citep{sun2025geovla}, VLA-4D~\citep{zhou2025vla4d}, 3D-CAVLA~\citep{bhat20253d}, QDepth-VLA~\citep{li2025qdepthvla}, and Spatial Forcing~\citep{li2025spatialforcing}.
LingBot-VA~\citep{li2026lingbotva} attains the highest average score, but uses a larger autoregressive video-action pipeline during control; GaussianDream instead uses prediction only as training-time supervision and keeps inference prefix-based.

\begin{table*}[t]
    \centering
    \small
    \setlength{\tabcolsep}{8pt}
    \begin{tabular}{lccccc}
        \toprule
        Method & Spatial & Object & Goal & Long & Average \\
        \midrule
        $\pi_0$~\citep{black2024pi0} & 96.8 & 98.8 & 95.8 & 85.2 & 94.1 \\
        $\pi_{0.5}$~\citep{physicalintelligence2025pi05} & 97.8 & 98.8 & 97.6 & 92.4 & 96.7 \\
        GeoPredict~\citep{qian2025geopredict} & 98.0 & 98.2 & 95.7 & 94.0 & 96.5 \\
        QDepth-VLA~\citep{li2025qdepthvla} & 97.6 & 96.6 & 95.2 & 90.0 & 94.9 \\
        LingBot-VA~\citep{li2026lingbotva} & 98.5 & 99.6 & 97.2 & \textbf{98.5} & \textbf{98.5} \\
        GeoVLA~\citep{sun2025geovla} & 98.4 & 99.0 & 96.6 & 96.6 & 97.7 \\
        VLA-4D~\citep{zhou2025vla4d} & 97.9 & 98.6 & 97.8 & 94.8 & 97.4 \\
        3D-CAVLA~\citep{bhat20253d} & 98.2 & \textbf{99.8} & 98.2 & 96.1 & 98.1 \\
        Spatial Forcing (PyTorch)~\citep{li2025spatialforcing} & 98.6 & 98.4 & 98.2 & 95.4 & 97.6 \\
        \textbf{GaussianDream} & \textbf{99.0} & 99.6 & \textbf{99.0} & 96.0 & 98.4 \\
        \bottomrule
    \end{tabular}
    \caption{
    Success rates (\%) on LIBERO~\citep{liu2023libero}. 
    GaussianDream achieves the best Spatial and Goal scores and 98.4\% average success; Spatial Forcing uses its PyTorch implementation.
    }
    \label{tab:libero_main_results}
\end{table*}

\paragraph{RoboCasa.}
GaussianDream shows its clearest advantage on spatially precise pick-and-place tasks in RoboCasa~\citep{nasiriany2024robocasa}.
As shown in Tab.~\ref{tab:robocasa_main_results}, it achieves the best average success rate of 54.8\% and the strongest Pick\&Place performance.
This suggests that reconstruction and prediction improve localization-sensitive manipulation.
Although GeoPredict~\citep{qian2025geopredict} leads on Doors/Drawers and Others, GaussianDream outperforms Being-H0.5~\citep{luo2026being} and achieves the best average.

\begin{table*}[t]
    \centering
    \small
    \setlength{\tabcolsep}{8pt}
    \begin{tabular}{lcccc}
        \toprule
        Method & Pick\&Place & Doors/Drawers & Others & Average \\
        \midrule
        $\pi_0$~\citep{black2024pi0} & 14.0 & 53.1 & 58.5 & 42.4 \\
        $\pi_{0.5}$~\citep{physicalintelligence2025pi05} & 36.0 & 46.5 & 39.5 & 40.1 \\
        BC-Transformer~\citep{nasiriany2024robocasa} & 3.8 & 46.7 & 38.0 & 28.8 \\
        GWM~\citep{lu2025gwm} & 14.8 & 54.3 & 49.8 & 39.3 \\
        GeoPredict~\citep{qian2025geopredict} & 22.7 & \textbf{75.1} & \textbf{62.4} & 52.4 \\
        Being-H0.5~\citep{luo2026being} & 36.0 & 71.7 & 57.6 & 53.9 \\
        \textbf{GaussianDream} & \textbf{43.8} & 66.3 & 54.4 & \textbf{54.8} \\
        \bottomrule
    \end{tabular}
    \caption{
    Success rates (\%) on RoboCasa Human-50~\citep{nasiriany2024robocasa}. 
    GaussianDream achieves the best average result and strongest Pick\&Place performance.
    }
    \label{tab:robocasa_main_results}
\end{table*}

\paragraph{Real robot.}
GaussianDream also improves physical execution.
As shown in Tab.~\ref{tab:real_robot_main_results}, it raises the average success rate from 34.4\% to 50.0\% over $\pi_{0.5}$~\citep{physicalintelligence2025pi05}, with gains across all scenario groups.
The largest gains appear in spatial-relation and long-horizon scenarios, where the GaussianDream prefix provides additional geometric context for action generation.

\begin{table*}[t]
    \centering
    \begin{minipage}[t]{0.48\textwidth}
        \centering
        \small
        \resizebox{\linewidth}{!}{%
        \begin{tabular}{lccccc}
            \toprule
            Method & Scene-A & Scene-B & Scene-C & Scene-D & Average \\
            \midrule
            $\pi_{0.5}$ & 42.5 & 50.0 & 25.0 & 20.0 & 34.4 \\
            \textbf{GaussianDream} & \textbf{55.0} & \textbf{70.0} & \textbf{35.0} & \textbf{40.0} & \textbf{50.0} \\
            \bottomrule
        \end{tabular}%
        }
        \caption{
        Real-robot success rates (\%). GaussianDream improves physical execution from 34.4\% to 50.0\%.
        }
        \label{tab:real_robot_main_results}
    \end{minipage}
    \hfill
    \begin{minipage}[t]{0.48\textwidth}
        \centering
        \small
        \resizebox{\linewidth}{!}{%
        \begin{tabular}{lccccc}
            \toprule
            Current Reconstruction & \ding{51} & \ding{51} & \ding{51} & \ding{51} & \ding{51} \\
            Future Prediction      & \ding{55} & \ding{55} & \ding{51} & \ding{51} & \ding{51} \\
            Rendering Branch       & \ding{55} & \ding{51} & \ding{55} & \ding{51} & \ding{51} \\
            Depth Branch           & \ding{55} & \ding{51} & \ding{51} & \ding{55} & \ding{51} \\
            \midrule
            Avg. on LIBERO~\citep{liu2023libero} & 97.0 & 97.3 & 97.5 & 97.2 & \textbf{98.4} \\
            \bottomrule
        \end{tabular}%
        }
        \caption{
        Ablation on LIBERO. 
        Reconstruction, prediction, rendering, and depth supervision provide complementary gains.
        }
        \label{tab:ablation_libero}
    \end{minipage}
\end{table*}

\subsection{Qualitative Analysis}

The qualitative results further validate the core design of GaussianDream: current reconstruction provides explicit spatial grounding, while future prediction captures short-horizon geometric evolution.
Fig.~\ref{fig:quantitative_analysis_vis} presents depth renderings for both current reconstruction and future prediction.
The reconstructed results preserve accurate object layouts in the current frame, while the predicted future states remain temporally coherent across consecutive steps.
Moreover, GaussianDream effectively models object interactions and coordinated motions, as evidenced by the visualized predictions.
These observations suggest that GaussianDream learns structured spatial-temporal representations from consecutive observation frames.
\subsection{Ablation Studies}

\paragraph{Component analysis.}
The ablation confirms that GaussianDream benefits from both current reconstruction and future prediction.
As shown in Tab.~\ref{tab:ablation_libero}, current reconstruction alone achieves 97.0\%, indicating that reconstructing the observation into a Gaussian state provides a strong spatial prior.
Adding future prediction improves performance to 97.5\%, showing that short-horizon state-change supervision contributes beyond current-frame grounding.
This supports the design in which the static head builds a spatial template and the dynamic head learns interaction-induced changes.

\paragraph{Supervision analysis.}
Rendering and depth losses provide complementary dense supervision.
Rendering improves the reconstruction-only variant from 97.0\% to 97.3\%, suggesting that image-level consistency helps align predicted Gaussians with observations.
When future prediction and rendering are retained but depth is removed, performance drops to 97.2\%, indicating that RGB consistency alone cannot fully constrain metric geometry.
The full model reaches 98.4\%, confirming the complementary roles of reconstruction, prediction, rendering, and depth.

%% file: sec/5._conclusion.tex
\section{Conclusion}
\label{sec:conclu}

We presented \textbf{GaussianDream}, a unified feed-forward 3D Gaussian world-model plug-in for language-conditioned robotic manipulation.
GaussianDream addresses three VLA limitations by grounding actions in reconstructed 3D Gaussian states, mining dense RGB/depth/scene-flow supervision from robot trajectories, and learning horizon-conditioned future prediction for short-horizon environment emulation.
Its asymmetric design uses full Gaussian reconstruction and prediction as training supervision, but discards all auxiliary decoding heads at deployment and retains only a compact GaussianDream prefix for action generation.
Experiments on LIBERO, RoboCasa Human-50, and real-robot tasks demonstrate strong performance, while ablations validate the complementary roles of reconstruction, prediction, rendering, and depth supervision.
Overall, GaussianDream shows that structured 3D Gaussian reconstruction and future emulation provide dense training-time supervision for physically grounded policies without sacrificing efficient closed-loop inference.

%% file: sec/6._appendix.tex
\appendix
\section*{Appendix}

\subsection*{A. Notation Summary}

\begin{table}[h]
    \centering
    \small
    \setlength{\tabcolsep}{6pt}
    \renewcommand{\arraystretch}{1.05}
    \begin{tabular*}{\textwidth}{@{\extracolsep{\fill}}llp{0.56\textwidth}@{}}
        \toprule
        Symbol & Name & Meaning \\
        \midrule
        $\mathbf{o}_t$, $\mathbf{o}_{t-K:t}$ & Observations & Current multi-view RGB observation and temporal observation window. \\
        $\mathbf{l}$, $\mathbf{s}_t$, $\mathbf{a}_t$ & Language, state, action & Language instruction, robot state, and predicted action chunk. \\
        $\pi_{\theta}$ & Action policy & Base continuous-action policy conditioned on the GaussianDream prefix. \\
        $\mathbf{Z}_t^{\mathrm{GD}}$ & GaussianDream prefix & Learned prefix tokens used for current reconstruction, future prediction, and action conditioning. \\
        $\mathbf{Q}_{\mathrm{GD}}$, $\mathcal{F}_{\omega}$ & GaussianDream queries/encoder & Learnable queries and temporal encoder used to build $\mathbf{Z}_t^{\mathrm{GD}}$ from multi-frame visual evidence. \\
        TGE Module & Temporal Gaussian Evolution Module & Module that aggregates multi-frame 3D-aware visual tokens and GaussianDream queries. \\
        $\mathcal{B}_{\mathrm{G}}$ & Gaussian decoder backbone & Decoder backbone that maps $\mathbf{Z}_t^{\mathrm{GD}}$ to dense Gaussian feature maps. \\
        $\mathbf{F}_t^{\mathrm{G}}$ & Gaussian feature map & Shared dense feature map produced by the Gaussian decoder. \\
        $\mathcal{R}_{\phi}$ & Reconstruction decoder & Auxiliary decoder for current Gaussian reconstruction. \\
        $\mathcal{D}_{\psi}$ & Prediction decoder & Auxiliary decoder for future Gaussian prediction. \\
        \midrule
        $\mathcal{G}_t$ & Current Gaussian state & Reconstructed current 3D Gaussian state. \\
        $\hat{\mathcal{G}}_{t+\Delta}$ & Future Gaussian state & Predicted Gaussian state at future horizon $\Delta$. \\
        $\mathbf{D}_t$, $\Theta_t$ & Depth / attributes & Current depth map and Gaussian attribute map. \\
        $\Theta_t^{\mathrm{geo}}$, $\Theta_t^{\mathrm{app}}$ & Geometry / appearance attributes & Gaussian geometry attributes and spherical-harmonics appearance coefficients. \\
        $\boldsymbol{\mu}_i^t$, $\theta_i^t$ & Gaussian primitive & Center and non-positional attributes of the $i$-th current Gaussian. \\
        $\hat{\boldsymbol{\mu}}_i^{\,t+\Delta}$ & Future center & Predicted future center of the $i$-th Gaussian. \\
        $\boldsymbol{\nu}_t^{(\Delta)}$, $\Delta\mathbf{X}_t^{(\Delta)}$ & Motion prediction & Horizon-conditioned velocity and center displacement. \\
        $\mathbf{e}_{\Delta}$, $\alpha_{\Delta}$ & Horizon terms & Horizon embedding and time-scale factor. \\
        \bottomrule
    \end{tabular*}
    \caption{Core notation for GaussianDream and its Gaussian reconstruction/prediction modules.}
    \label{tab:notation_core}
\end{table}

\begin{table}[h]
    \centering
    \small
    \setlength{\tabcolsep}{6pt}
    \renewcommand{\arraystretch}{1.05}
    \begin{tabular*}{\textwidth}{@{\extracolsep{\fill}}llp{0.56\textwidth}@{}}
        \toprule
        Symbol & Name & Meaning \\
        \midrule
        $\mathcal{H}$, $w_{\Delta}$ & Horizons & Future horizon set and horizon-specific loss weight. \\
        $\mathcal{L}_{\mathrm{cur}}$ & Current loss & Current depth and rendering reconstruction loss. \\
        $\mathcal{L}_{\mathrm{fut}}$ & Future loss & Future depth, rendering, and pseudo scene-flow supervision loss. \\
        $\mathcal{L}_{\mathrm{GD}}$ & GaussianDream objective & Combined current reconstruction and future prediction objective. \\
        $\mathbf{c}_t$ & Policy context & Action-conditioning context with visual observation, language, robot state, and GaussianDream prefix. \\
        $\tau$, $\boldsymbol{\epsilon}$ & Flow variables & Flow-matching time and Gaussian noise. \\
        $\mathbf{v}_{\theta}$ & Flow field & Policy-predicted velocity field in the action objective. \\
        $\mathcal{L}_{\mathrm{act}}$ & Action loss & Flow-matching action loss from the base $\pi_{0.5}$ policy. \\
        \bottomrule
    \end{tabular*}
    \caption{Training notation used in the GaussianDream and action objectives.}
    \label{tab:notation_training}
\end{table}

\subsection*{B. Architecture Details}

This section provides implementation details for the GaussianDream plug-in described in Sec.~\ref{sec:method}. 
The architecture is designed to keep the online policy lightweight while enabling training-time 3D reconstruction and future emulation. 
Concretely, the GaussianDream token path extracts temporal 3D-aware evidence into a compact prefix, and the Gaussian decoder uses this prefix only during training to produce current and future Gaussian states for dense supervision.

The GaussianDream token path uses a dense 32$\times$32 query grid, corresponding to 1024 learnable GaussianDream queries. 
Each query is first projected to 512 channels and processed by the TGE Module. 
Each TGE Module block performs self-attention over the joint query--patch tokens within each frame, followed by temporal attention over the same token slot across frames. 
We use 8 attention heads and a 4$\times$ expansion MLP. 
The temporal visual tokens are obtained from VGGT~\citep{wang2025vggt} feature taps using 32$\times$32 adaptive average pooling and linear projection to 512 channels. 
The TGE Module output at the latest frame is projected back to the VLM width and appended to the multimodal prefix as the GaussianDream token segment. 
After VLM prefix processing, this segment becomes $\mathbf{Z}_t^{\mathrm{GD}}$.

The Gaussian decoder reshapes $\mathbf{Z}_t^{\mathrm{GD}}$ into a canonical 32$\times$32 token grid and applies a shared token-to-feature backbone. 
The backbone uses three transposed-convolution upsampling blocks with kernel size 4, stride 2, and padding 1, progressively lifting the token grid to a 256$\times$256 feature map while reducing channels from the VLM width to 512, 256, and finally 128. 
Each upsampling block contains GroupNorm, GELU, a 3$\times$3 convolution, and a bilinear residual skip. 
The resulting 128-channel feature map is further refined by DPT-style feature fusion blocks with 3$\times$3 and 1$\times$1 convolutions and residual connections from intermediate resolutions.

The current reconstruction head operates on the shared 256$\times$256 feature map. 
Its geometry branch predicts 8 channels with a 3$\times$3 convolution: quaternion rotation (4), scale (3), and opacity (1). 
A separate depth branch uses two 3$\times$3 convolutional layers with GroupNorm and GELU, followed by a final 3$\times$3 convolution to predict one depth channel. 
The appearance branch optionally fuses the RGB image through a 7$\times$7 convolution and predicts 9 spherical-harmonic appearance channels with a 3$\times$3 convolution. 
Depth is unprojected with camera intrinsics to form Gaussian centers, while the remaining channels provide Gaussian attributes.

The future prediction head reuses the same shared 256$\times$256 feature map. 
A horizon embedding is added after a 3$\times$3 projection to 128 channels, followed by a residual block with two 3$\times$3 convolutions and GroupNorm. 
The velocity output head then applies a 3$\times$3 convolution, GroupNorm, GELU, and a final 1$\times$1 convolution to predict a 3-channel center-velocity map. 
The velocity is passed through \texttt{tanh} and scaled, then multiplied by the horizon time factor to update only Gaussian centers; scale, opacity, appearance, and rotation are copied from the current Gaussian template. 
This design concentrates future prediction on short-horizon geometric change while preserving the stability of the reconstructed current Gaussian state.
\paragraph{Training and Evaluation Details.}
We train our model for 60K optimization steps with a global batch size of 24.
We use AdamW as the optimizer with a peak learning rate of $5\times10^{-5}$ and a cosine learning-rate schedule with 10K warmup steps.
Gradient clipping is applied with a maximum norm of 1.0.
We use an exponential moving average (EMA) of model parameters with a decay rate of 0.999.
All experiments are conducted with mixed-precision training on NVIDIA A100 GPUs.

During evaluation on LIBERO~\citep{liu2023libero} and RoboCasa~\citep{nasiriany2024robocasa}, each task is executed for 50 consecutive trials following the standard benchmark protocols.
The reported success rates are averaged across all evaluation trials.

\subsection*{C. Pseudo Depth and Scene-Flow Preparation}

GaussianDream uses pseudo geometric supervision to train current Gaussian reconstruction and future Gaussian prediction.
For each demonstration frame, we first generate a dense pseudo depth map from the agent-view RGB observation using Depth Anything V2~\citep{yang2024depth}.
The predicted depth is resized back to the original image resolution and stored together with the episode data.
These pseudo geometric targets are used only during training.
They allow GaussianDream to transform ordinary demonstration sequences into dense spatial-temporal supervision, instead of relying solely on sparse action labels.

To supervise future prediction, we construct pseudo 3D scene-flow sidecars from temporally adjacent frames.
Given two consecutive RGB observations $(\mathbf{o}_t, \mathbf{o}_{t+1})$, we estimate a dense 2D optical flow field $\mathbf{f}_{t\rightarrow t+1}$ using a RAFT~\citep{teed2020raft} backend by default, with Farneback flow as a lightweight fallback.
The 2D flow warps each pixel $(u,v)$ in frame $t$ to $(u',v')$ in frame $t{+}1$:
\begin{equation}
    (u',v') = (u,v) + \mathbf{f}_{t\rightarrow t+1}(u,v).
\end{equation}

We then bilinearly sample the future depth map at $(u',v')$ and back-project both frames with the camera intrinsics $\mathbf{K}$:
\begin{equation}
    \mathbf{x}_t = \Pi^{-1}(u,v,\mathbf{D}_t(u,v);\mathbf{K}), \qquad
    \mathbf{x}_{t+1} = \Pi^{-1}(u',v',\mathbf{D}_{t+1}(u',v');\mathbf{K}).
\end{equation}

The pseudo 3D flow target is computed as the difference between the two back-projected points:
\begin{equation}
    \mathbf{F}^{3D}_{t\rightarrow t+1}(u,v)
    =
    \mathbf{x}_{t+1}
    -
    \mathbf{x}_t.
\end{equation}

We also compute a validity mask that removes correspondences warped outside the image, pixels with invalid depth, and pixels outside the configured depth range.
Invalid flow vectors are set to zero and excluded from the future prediction loss.
Each sidecar stores the 2D flow, pseudo 3D flow, validity mask, camera intrinsics, camera name, and metadata such as episode and task indices.

\begin{figure*}[!t]
    \centering
    \includegraphics[width=\textwidth]{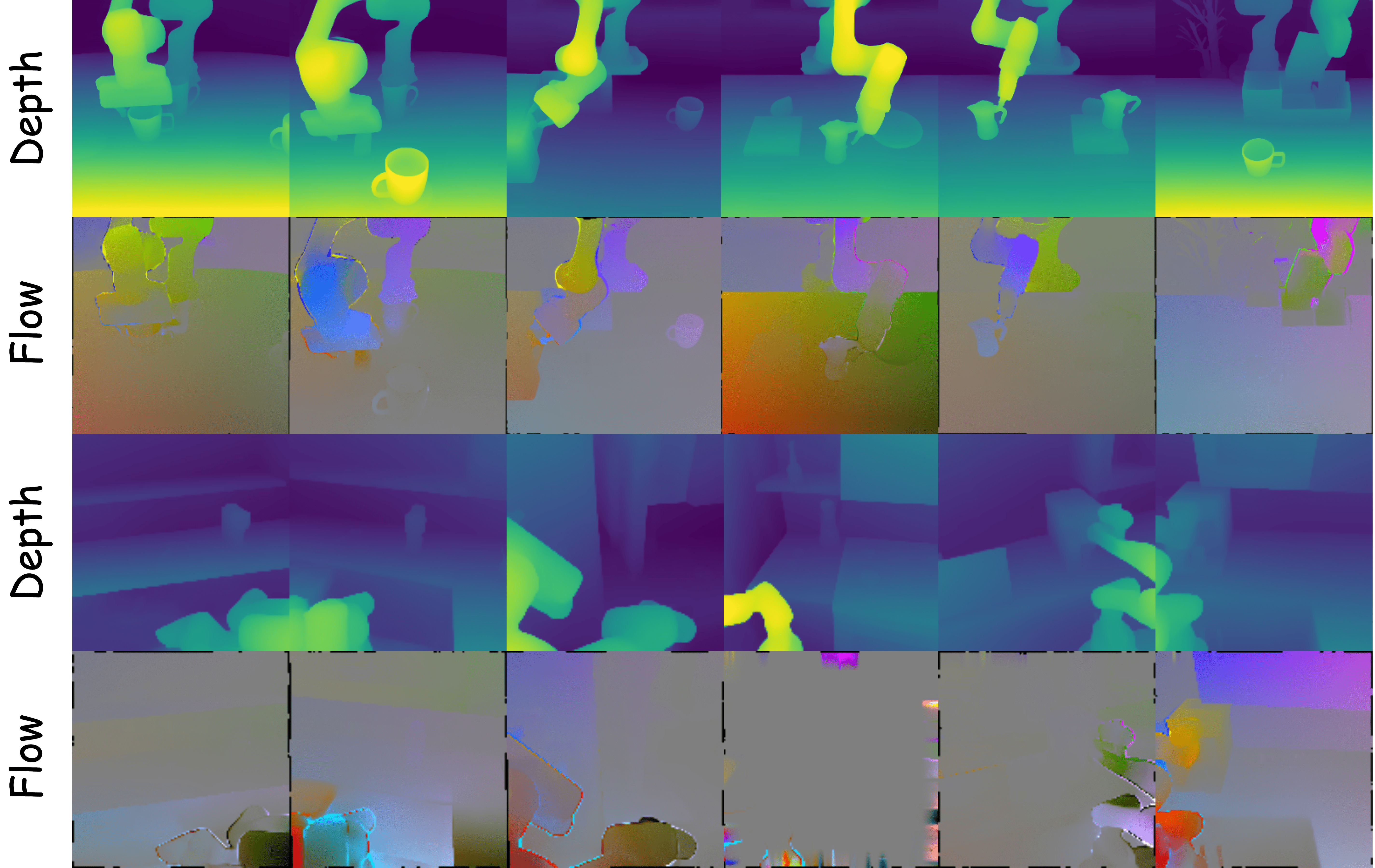}
    \caption{
    Visualization of pseudo depth and scene-flow preparation on LIBERO~\citep{liu2023libero} and RoboCasa~\citep{nasiriany2024robocasa} demonstrations.
    The first two rows show LIBERO depth maps and flow fields, while the last two rows show RoboCasa depth maps and flow fields.
    These visualizations illustrate the pseudo geometric supervision used to construct 3D scene-flow targets for future Gaussian prediction.
    }
    \label{fig:appendix_pseudo_depth_flow}
\end{figure*}

\FloatBarrier

\subsection*{D. Loss Details}

We provide additional implementation details for the GaussianDream objectives introduced.
The current reconstruction objective consists of depth and RGB rendering supervision:
\begin{equation}
    \mathcal{L}_{\mathrm{cur}}
    =
    \lambda_{\mathrm{cur}}^{\mathrm{depth}}
    \mathcal{L}_{\mathrm{cur}}^{\mathrm{depth}}
    +
    \lambda_{\mathrm{cur}}^{\mathrm{render}}
    \mathcal{L}_{\mathrm{cur}}^{\mathrm{render}}.
\end{equation}
The depth loss aligns the reconstructed Gaussian depth with pseudo or simulator-provided depth targets, while the rendering loss supervises RGB renderings against the corresponding observation frames.

The future prediction objective applies depth, rendering, and pseudo 3D scene-flow supervision over the prediction horizon set $\mathcal{H}$:
\begin{equation}
    \mathcal{L}_{\mathrm{fut}}
    =
    \sum_{\Delta\in\mathcal{H}}w_{\Delta}
    \left(
    \lambda_{\mathrm{depth}}
    \mathcal{L}_{\mathrm{depth}}^{(\Delta)}
    +
    \lambda_{\mathrm{render}}
    \mathcal{L}_{\mathrm{render}}^{(\Delta)}
    +
    \lambda_{\mathrm{flow}}
    \mathcal{L}_{\mathrm{flow}}^{(\Delta)}
    \right).
\end{equation}

The future depth and rendering losses maintain geometric and visual consistency between predicted future Gaussians and future observations.
The flow objective directly supervises predicted Gaussian center displacements using pseudo 3D scene flow.
With a validity mask, the flow loss is defined as
\begin{equation}
    \mathcal{L}_{\mathrm{flow}}^{(\Delta)}
    =
    \frac{
    \sum_i
    \mathbf{M}_i^{(\Delta)}
    \left\|
    \Delta\mathbf{x}_i^{(\Delta)}
    -
    \mathbf{F}_{i}^{3D,(\Delta)}
    \right\|_1
    }{
    \sum_i
    \mathbf{M}_i^{(\Delta)}
    +
    \epsilon
    },
\end{equation}
where $\mathbf{M}_i^{(\Delta)}$ denotes the validity mask and $\mathbf{F}_{i}^{3D,(\Delta)}$ denotes the pseudo 3D scene-flow target sampled at the corresponding Gaussian or pixel location.

The complete GaussianDream objective is
\begin{equation}
    \mathcal{L}_{\mathrm{GD}}
    =
    \mathcal{L}_{\mathrm{cur}}
    +
    \mathcal{L}_{\mathrm{fut}}.
\end{equation}

During policy learning, the action objective is jointly optimized with the GaussianDream objective:
\begin{equation}
    \mathcal{L}
    =
    \mathcal{L}_{\mathrm{act}}
    +
    \lambda_{\mathrm{GD}}
    \mathcal{L}_{\mathrm{GD}}.
\end{equation}

In practice, the prediction horizon is gradually increased during training for stable optimization, and all loss terms are computed only during training.
At inference time, all reconstruction and prediction heads are removed, while only the compact GaussianDream prefix is retained for policy conditioning.

\FloatBarrier

\subsection*{E. Real-Robot Hardware Setup}

Our real-robot platform contains two robot arms with distinct roles. 
The \emph{leader arm} is used only during teleoperation-based data collection, where a human operator manually controls the arm to generate demonstrations. 
The \emph{follower arm} is the execution arm: it follows the leader arm during demonstration collection and is controlled by the learned policy during autonomous evaluation. 
This leader--follower design separates demonstration acquisition from policy execution and allows us to evaluate the learned controller on the same physical embodiment used for data collection.

The policy receives language instructions, robot state, and visual observations from two RGB cameras. 
The agent-view camera captures the overall workspace, object arrangement, and spatial relations, while the wrist-mounted camera provides close-range observations around the gripper. 
During evaluation, GaussianDream encodes the temporal visual buffer into GaussianDream prefix tokens and conditions the action policy on this prefix. 
The auxiliary rendering, depth, and velocity heads used during training are removed at deployment, so the policy directly outputs action chunks for the follower arm without test-time rendering or future-frame prediction.

\begin{figure*}[!t]
    \centering
    \includegraphics[width=0.92\textwidth]{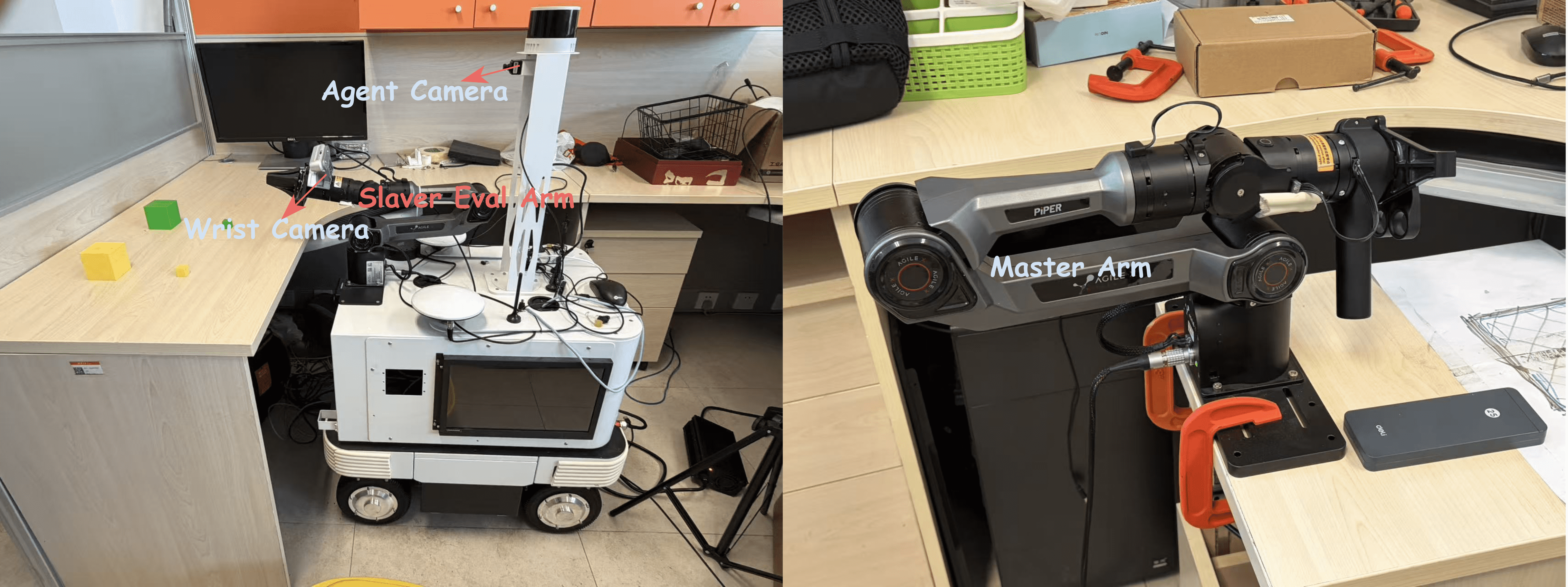}
    \caption{
    Real-robot hardware setup. 
    The evaluation platform uses a policy-controlled follower arm mounted on a mobile base, together with an agent-view camera and a wrist-mounted camera for visual observation. 
    During teleoperation-based demonstration collection, a separate leader arm is manually operated by the user, and its motions are transferred to the follower arm. 
    During autonomous evaluation, GaussianDream directly controls the follower arm from language instructions and visual observations.
    }
    \label{fig:hardware_setup}
\end{figure*}

\begin{figure*}[!t]
    \centering
    \includegraphics[width=\textwidth]{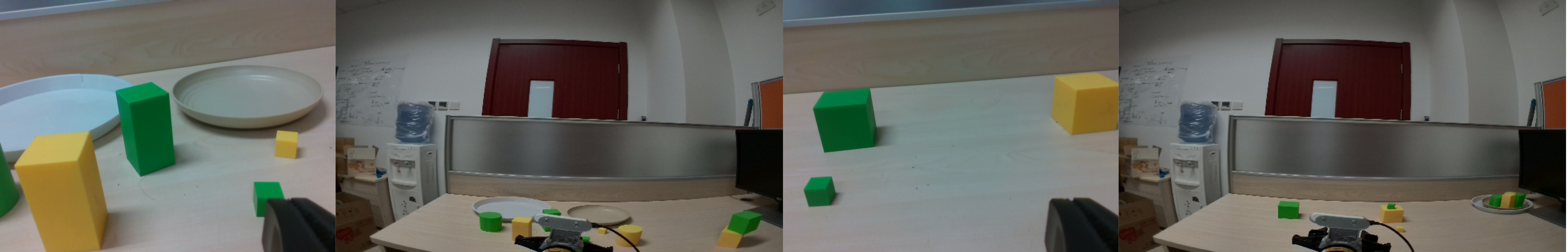}
    \caption{
    Real-robot visual observations from the agent-view and wrist-mounted cameras. 
    The agent-view camera provides a global view of the workspace and object layout, while the wrist-mounted camera provides local visual feedback near the end effector. 
    These complementary views are used by GaussianDream during physical robot evaluation.
    }
    \label{fig:appendix_piper_agent_wrist_vis}
\end{figure*}

\FloatBarrier

\subsection*{F. Real-Robot Supplementary Results}

\begin{figure*}[!t]
    \centering
    \begin{minipage}[t]{0.48\textwidth}
        \centering
        \includegraphics[width=\linewidth]{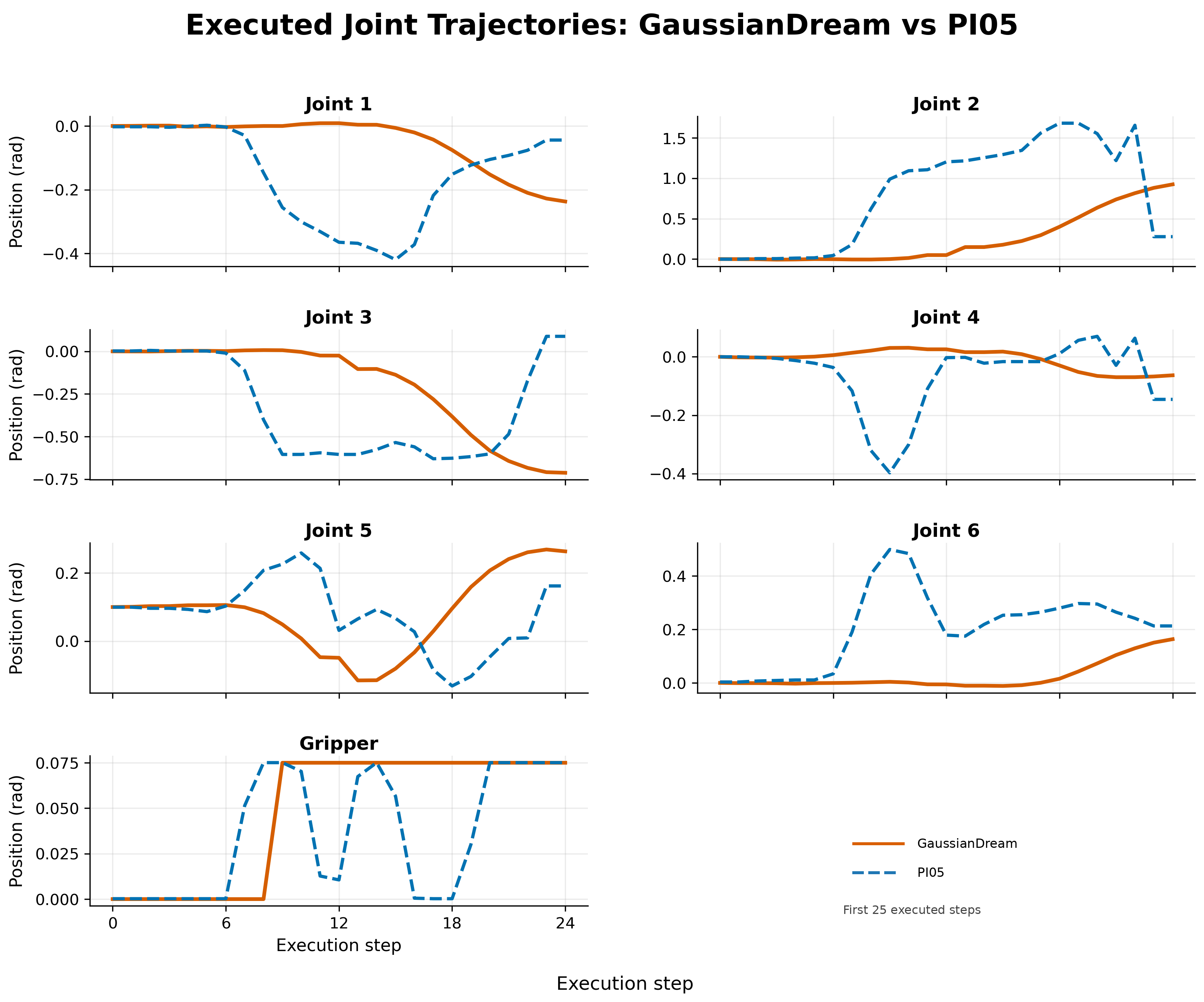}
    \end{minipage}
    \hfill
    \begin{minipage}[t]{0.48\textwidth}
        \centering
        \includegraphics[width=\linewidth]{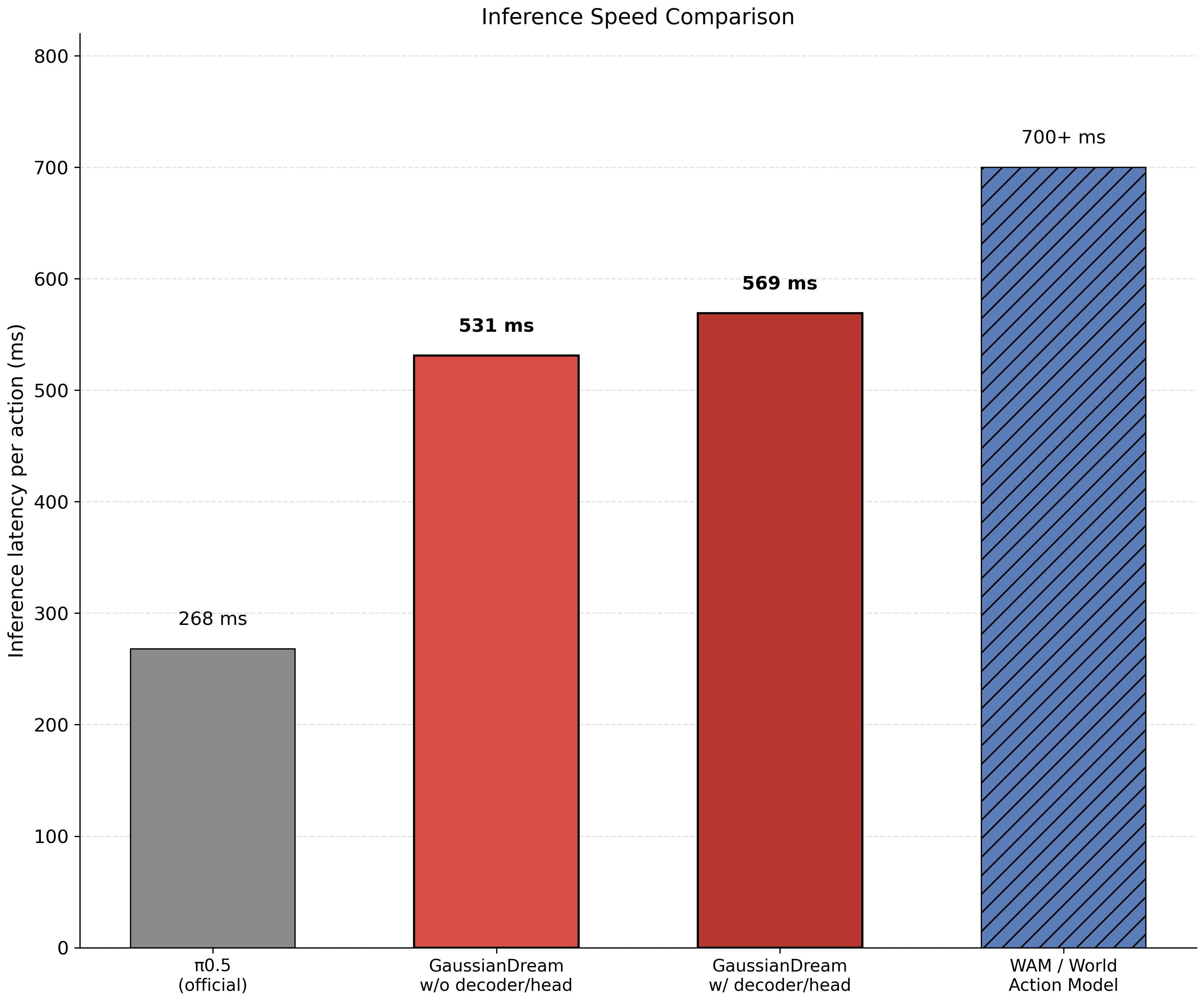}
    \end{minipage}
    \caption{
    Real-robot smoothness and inference-speed analysis.
    Left: execution smoothness comparison between GaussianDream and the $\pi_{0.5}$~\citep{physicalintelligence2025pi05} baseline.
    Right: per-action-chunk inference latency comparison across GaussianDream variants and the WAM / World Action Model baseline.
    }
    \label{fig:appendix_real_robot_smooth_speed}
\end{figure*}

We provide additional real-robot analysis to complement the physical success rates reported in the main paper.
Fig.~\ref{fig:appendix_real_robot_smooth_speed} shows that GaussianDream reduces abrupt trajectory changes compared with the $\pi_{0.5}$~\citep{physicalintelligence2025pi05} baseline, suggesting that the learned GaussianDream prefix provides useful spatial context for action generation.
The same figure also highlights the efficiency of the asymmetric design: at deployment, the auxiliary Gaussian decoder and prediction heads are removed, yielding 531 ms per action chunk, while the diagnostic configuration that retains the decoder/head remains at 569 ms per action chunk.
Both variants are faster than the WAM / World Action Model baseline, which requires more than 700 ms per action chunk, supporting the practicality of GaussianDream for online robot control.

\FloatBarrier

\subsection*{G. Additional Qualitative Visualization}

These visualizations complement the quantitative results in the main paper. 
The current reconstruction branch recovers coarse object layout and depth structure from the observed frame, showing that the GaussianDream prefix can be decoded into a spatially grounded 3D representation. 
The future prediction branch produces temporally coherent short-horizon Gaussian changes, indicating that the model learns how the reconstructed Gaussian state evolves across nearby future frames. 
This qualitative behavior supports the central motivation of GaussianDream: current reconstruction mitigates spatial and geometric underspecification, future prediction introduces short-horizon environment emulation, and depth/rendering supervision converts continuous robot observation frames into dense spatial-temporal learning signals.

\begin{figure*}[!t]
    \centering
    \includegraphics[width=0.92\textwidth]{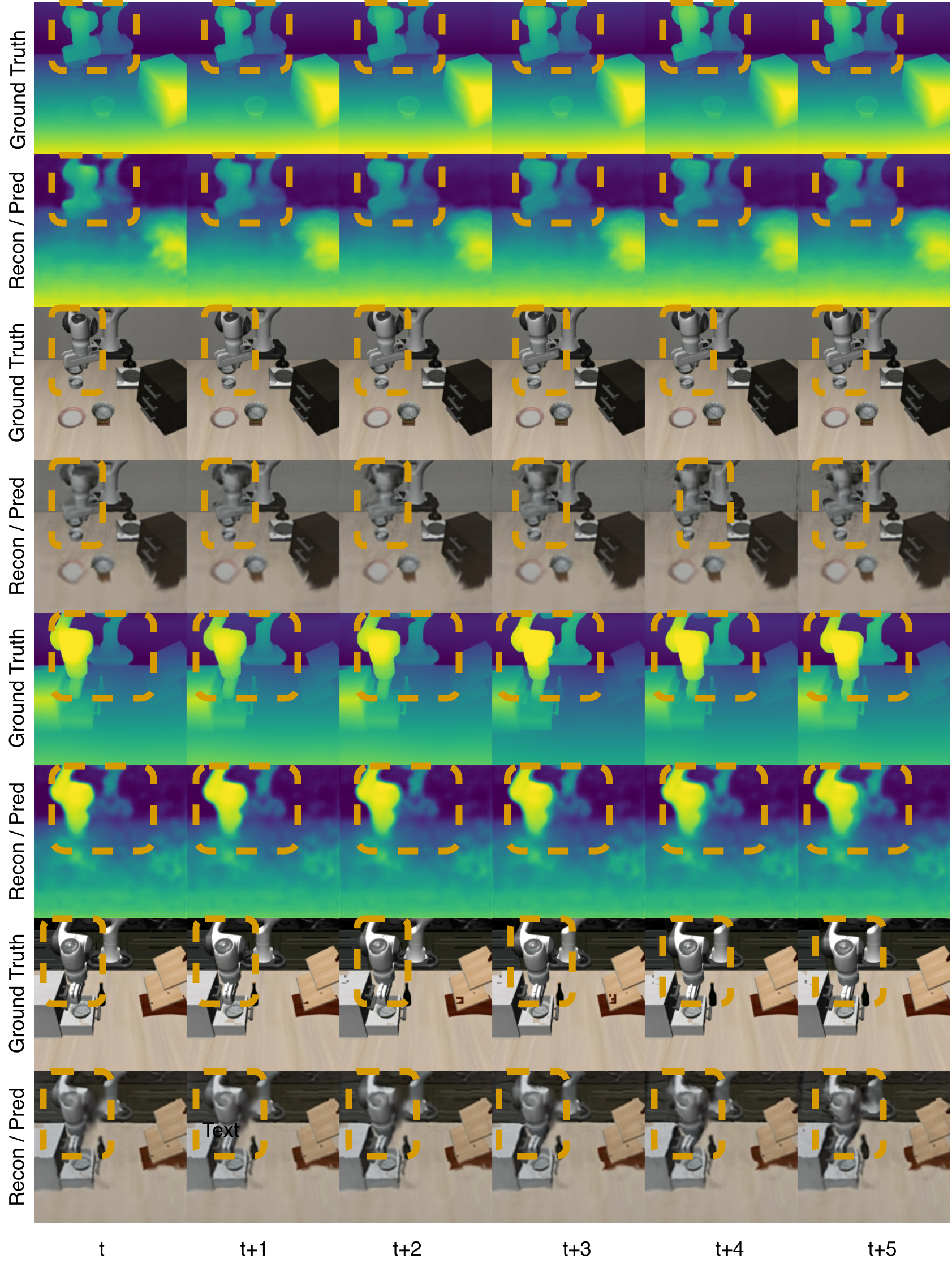}
    \caption{
    Additional qualitative visualization of current Gaussian reconstruction and future Gaussian prediction. 
    Each example compares the observed frame and pseudo depth target with the reconstructed current geometry and predicted future Gaussian states. 
    The visualization illustrates that GaussianDream captures both current scene layout and short-horizon geometric changes.
    }
    \label{fig:appendix_qualitative_geometry}
\end{figure*}

\FloatBarrier

%% file: main.bib
@String(ICCV= {Int. Conf. Comput. Vis.})

@String(ICCV  = {ICCV})

@article{kim2024openvla,
    author  = {Moo Jin Kim and Karl Pertsch and Siddharth Karamcheti and Ted Xiao and Ashwin Balakrishna
  and Suraj Nair and Rafael Rafailov and Ethan Foster and Grace Lam and Pannag Sanketi and Quan Vuong and
  Thomas Kollar and Benjamin Burchfiel and Russ Tedrake and Dorsa Sadigh and Sergey Levine and Percy Liang
  and Chelsea Finn},
    title   = {{OpenVLA}: An Open-Source Vision-Language-Action Model},
    journal = {CoRR},
    volume  = {abs/2406.09246},
    year    = {2024},
    url     = {https://arxiv.org/abs/2406.09246}
  }

@article{black2024pi0,
    author  = {Kevin Black and Noah Brown and Danny Driess and Adnan Esmail and Michael Equi and Chelsea
  Finn and Niccolo Fusai and Lachy Groom and Karol Hausman and Brian Ichter and Szymon Jakubczak and Tim
  Jones and Liyiming Ke and Sergey Levine and Adrian Li-Bell and Mohith Mothukuri and Suraj Nair and Karl
  Pertsch and Lucy Xiaoyang Shi and James Tanner and Quan Vuong and Anna Walling and Haohuan Wang and Ury
  Zhilinsky},
    title   = {{${\pi}_0$}: A Vision-Language-Action Flow Model for General Robot Control},
    journal = {CoRR},
    volume  = {abs/2410.24164},
    year    = {2024},
    url     = {https://arxiv.org/abs/2410.24164}
  }

@article{physicalintelligence2025pi05,
    author  = {{Physical Intelligence} and Kevin Black and Noah Brown and James Darpinian and Karan
  Dhabalia and Danny Driess and Adnan Esmail and Michael Equi and Chelsea Finn and Niccolo Fusai and Manuel
  Y. Galliker and Dibya Ghosh and Lachy Groom and Karol Hausman and Brian Ichter and Szymon Jakubczak and
  Tim Jones and Liyiming Ke and Devin LeBlanc and Sergey Levine and Adrian Li-Bell and Mohith Mothukuri and
  Suraj Nair and Karl Pertsch and others},
    title   = {{${\pi}_{0.5}$}: A Vision-Language-Action Model with Open-World Generalization},
    journal = {CoRR},
    volume  = {abs/2504.16054},
    year    = {2025},
    url     = {https://arxiv.org/abs/2504.16054}
  }

@article{li2025qdepthvla,
    author  = {Yixuan Li and Yuhui Chen and Mingcai Zhou and Haoran Li and Zhengtao Zhang and Dongbin
  Zhao},
    title   = {{QDepth-VLA}: Quantized Depth Prediction as Auxiliary Supervision for Vision-Language-Action
  Models},
    journal = {CoRR},
    volume  = {abs/2510.14836},
    year    = {2025},
    url     = {https://arxiv.org/abs/2510.14836}
  }

@article{song2026vga,
    author  = {Zijian Song and Qichang Li and Jiawei Zhou and Zhenlong Yuan and Tianshui Chen and Liang Lin
  and Guangrun Wang},
    title   = {Robotic Manipulation is Vision-to-Geometry Mapping ($f(v) \rightarrow G$): Vision-Geometry
  Backbones over Language and Video Models},
    journal = {CoRR},
    volume  = {abs/2604.12908},
    year    = {2026},
    url     = {https://arxiv.org/abs/2604.12908}
  }

@article{qian2025geopredict,
    author  = {Jingjing Qian and Boyao Han and Chen Shi and Lei Xiao and Long Yang and Shaoshuai Shi and Li
  Jiang},
    title   = {{GeoPredict}: Leveraging Predictive Kinematics and 3D Gaussian Geometry for Precise VLA
  Manipulation},
    journal = {CoRR},
    volume  = {abs/2512.16811},
    year    = {2025},
    url     = {https://arxiv.org/abs/2512.16811}
  }

@article{ye2026dreamzero,
    author  = {Seonghyeon Ye and Yunhao Ge and Kaiyuan Zheng and Shenyuan Gao and Sihyun Yu and George
  Kurian and Suneel Indupuru and You Liang Tan and Chuning Zhu and Jiannan Xiang and Ayaan Malik and
  Kyungmin Lee and William Liang and Nadun Ranawaka and Jiasheng Gu and Yinzhen Xu and Guanzhi Wang and
  Fengyuan Hu and Avnish Narayan and Johan Bjorck and Jing Wang and Gwanghyun Kim and Dantong Niu and
  Ruijie Zheng and Yuqi Xie and Jimmy Wu and Qi Wang and Ryan Julian and Danfei Xu and Yilun Du and Yevgen
  Chebotar and Scott Reed and Yuke Zhu and Linxi Fan and Joel Jang},
    title   = {{World Action Models are Zero-shot Policies}},
    journal = {CoRR},
    volume  = {abs/2602.15922},
    year    = {2026},
    url     = {https://arxiv.org/abs/2602.15922}
  }

@article{li2026lingbotva,
    author  = {Lin Li and Qihang Zhang and Yiming Luo and Shuai Yang and Ruilin Wang and Fei Han and
  Mingrui Yu and Zelin Gao and Nan Xue and Xing Zhu and Yujun Shen and Yinghao Xu},
    title   = {Causal World Modeling for Robot Control},
    journal = {CoRR},
    volume  = {abs/2601.21998},
    year    = {2026},
    url     = {https://arxiv.org/abs/2601.21998}
  }

@article{bi2025motus,
    author  = {Hongzhe Bi and Hengkai Tan and Shenghao Xie and Zeyuan Wang and Shuhe Huang and Haitian Liu
  and Ruowen Zhao and Yao Feng and Chendong Xiang and Yinze Rong and Hongyan Zhao and Hanyu Liu and
  Zhizhong Su and Lei Ma and Hang Su and Jun Zhu},
    title   = {{Motus}: A Unified Latent Action World Model},
    journal = {CoRR},
    volume  = {abs/2512.13030},
    year    = {2025},
    url     = {https://arxiv.org/abs/2512.13030}
  }

@article{kim2026cosmospolicy,
    author  = {Moo Jin Kim and Yihuai Gao and Tsung-Yi Lin and Yen-Chen Lin and Yunhao Ge and Grace Lam and
  Percy Liang and Shuran Song and Ming-Yu Liu and Chelsea Finn and Jinwei Gu},
    title   = {{Cosmos Policy}: Fine-Tuning Video Models for Visuomotor Control and Planning},
    journal = {CoRR},
    volume  = {abs/2601.16163},
    year    = {2026},
    url     = {https://arxiv.org/abs/2601.16163}
  }

@inproceedings{lu2025gwm,
    author    = {Guanxing Lu and Baoxiong Jia and Puhao Li and Yixin Chen and Ziwei Wang and Yansong Tang
  and Siyuan Huang},
    title     = {{GWM}: Towards Scalable Gaussian World Models for Robotic Manipulation},
    booktitle = ICCV,
    year      = {2025},
    pages     = {9263--9274},
    url       =
  {https://openaccess.thecvf.com/content/ICCV2025/html/Lu_GWM_Towards_Scalable_Gaussian_World_Models_for_Robotic_Manipulation_ICCV_2025_paper.html}
  }

@article{brohan2022rt1,
    author  = {Anthony Brohan and Noah Brown and Justice Carbajal and Yevgen Chebotar and others},
    title   = {{RT-1}: Robotics Transformer for Real-World Control at Scale},
    journal = {CoRR},
    volume  = {abs/2212.06817},
    year    = {2022},
    url     = {https://arxiv.org/abs/2212.06817}
  }

@article{brohan2023rt2,
    author  = {Anthony Brohan and Noah Brown and Justice Carbajal and Yevgen Chebotar and others},
    title   = {{RT-2}: Vision-Language-Action Models Transfer Web Knowledge to Robotic Control},
    journal = {CoRR},
    volume  = {abs/2307.15818},
    year    = {2023},
    url     = {https://arxiv.org/abs/2307.15818}
  }

@article{octo2024,
    author  = {{Octo Model Team} and Dibya Ghosh and Homer Walke and Karl Pertsch and Kevin Black and Oier
  Mees and others},
    title   = {Octo: An Open-Source Generalist Robot Policy},
    journal = {CoRR},
    volume  = {abs/2405.12213},
    year    = {2024},
    url     = {https://arxiv.org/abs/2405.12213}
  }

@article{pertsch2025fast,
    author  = {Karl Pertsch and Kyle Stachowicz and Brian Ichter and Danny Driess and Suraj Nair and Quan
  Vuong and Oier Mees and Chelsea Finn and Sergey Levine},
    title   = {{FAST}: Efficient Action Tokenization for Vision-Language-Action Models},
    journal = {CoRR},
    volume  = {abs/2501.09747},
    year    = {2025},
    url     = {https://arxiv.org/abs/2501.09747}
  }

@article{shukor2025smolvla,
    author  = {Mustafa Shukor and Dana Aubakirova and Francesco Capuano and Pepijn Kooijmans and Steven
  Palma and Adil Zouitine and Michel Aractingi and Caroline Pascal and Martino Russi and Andres Marafioti
  and Simon Alibert and Matthieu Cord and Thomas Wolf and Remi Cadene},
    title   = {SmolVLA: A Vision-Language-Action Model for Affordable and Efficient Robotics},
    journal = {CoRR},
    volume  = {abs/2506.01844},
    year    = {2025},
    url     = {https://arxiv.org/abs/2506.01844}
  }

@article{wen2024tinyvla,
    author  = {Junjie Wen and Yichen Zhu and Jinming Li and Minjie Zhu and Kun Wu and Zhiyuan Xu and Ning
  Liu and Ran Cheng and Chaomin Shen and Yaxin Peng and Feifei Feng and Jian Tang},
    title   = {TinyVLA: Towards Fast, Data-Efficient Vision-Language-Action Models for Robotic
  Manipulation},
    journal = {CoRR},
    volume  = {abs/2409.12514},
    year    = {2024},
    url     = {https://arxiv.org/abs/2409.12514}
  }

@article{li2025spatialforcing,
    author  = {Fuhao Li and Wenxuan Song and Han Zhao and Jingbo Wang and Pengxiang Ding and Donglin Wang
  and Long Zeng and Haoang Li},
    title   = {Spatial Forcing: Implicit Spatial Representation Alignment for Vision-Language-Action
  Model},
    journal = {CoRR},
    volume  = {abs/2510.12276},
    year    = {2025},
    url     = {https://arxiv.org/abs/2510.12276}
  }

@article{abouzeid2025geoawarevla,
    author  = {Ali Abouzeid and Malak Mansour and Zezhou Sun and Dezhen Song},
    title   = {{GeoAware-VLA}: Implicit Geometry Aware Vision-Language-Action Model},
    journal = {CoRR},
    volume  = {abs/2509.14117},
    year    = {2025},
    url     = {https://arxiv.org/abs/2509.14117}
  }

@article{sun2025geovla,
    author  = {Lin Sun and Bin Xie and Yingfei Liu and Hao Shi and Tiancai Wang and Jiale Cao},
    title   = {{GeoVLA}: Empowering 3D Representations in Vision-Language-Action Models},
    journal = {CoRR},
    volume  = {abs/2508.09071},
    year    = {2025},
    url     = {https://arxiv.org/abs/2508.09071}
  }

@article{deng2025stereovla,
    author  = {Shengliang Deng and Mi Yan and Yixin Zheng and Jiayi Su and Wenhao Zhang and Xiaoguang Zhao
  and Heming Cui and Zhizheng Zhang and He Wang},
    title   = {{StereoVLA}: Enhancing Vision-Language-Action Models with Stereo Vision},
    journal = {CoRR},
    volume  = {abs/2512.21970},
    year    = {2025},
    url     = {https://arxiv.org/abs/2512.21970}
  }

@article{zhou2025vla4d,
    author  = {Hanyu Zhou and Chuanhao Ma and Gim Hee Lee},
    title   = {{VLA-4D}: Embedding 4D Awareness into Vision-Language-Action Models for SpatioTemporally
  Coherent Robotic Manipulation},
    journal = {CoRR},
    volume  = {abs/2511.17199},
    year    = {2025},
    url     = {https://arxiv.org/abs/2511.17199}
  }

@article{ni2025swiftvla,
    author  = {Chaojun Ni and Cheng Chen and Xiaofeng Wang and Zheng Zhu and Wenzhao Zheng and Boyuan Wang
  and Tianrun Chen and Guosheng Zhao and Haoyun Li and Zhehao Dong and Qiang Zhang and Yun Ye and Yang Wang
  and Guan Huang and Wenjun Mei},
    title   = {{SwiftVLA}: Unlocking Spatiotemporal Dynamics for Lightweight VLA Models at Minimal
  Overhead},
    journal = {CoRR},
    volume  = {abs/2512.00903},
    year    = {2025},
    url     = {https://arxiv.org/abs/2512.00903}
  }

@inproceedings{wang2025vggt,
  title={Vggt: Visual geometry grounded transformer},
  author={Wang, Jianyuan and Chen, Minghao and Karaev, Nikita and Vedaldi, Andrea and Rupprecht, Christian and Novotny, David},
  booktitle={Proceedings of the Computer Vision and Pattern Recognition Conference},
  pages={5294--5306},
  year={2025}
}

@inproceedings{zhao2025vlas,
  title={Vlas: Vision-language-action model with speech instructions for customized robot manipulation},
  author={Zhao, Wei and Ding, Pengxiang and Min, Zhang and Gong, Zhefei and Bai, Shuanghao and Zhao, Han and Wang, Donglin},
  booktitle={International conference on learning representations},
  volume={2025},
  pages={51676--51693},
  year={2025}
}

@article{kerbl20233d,
  title={3d gaussian splatting for real-time radiance field rendering.},
  author={Kerbl, Bernhard and Kopanas, Georgios and Leimk{\"u}hler, Thomas and Drettakis, George and others},
  journal={ACM Trans. Graph.},
  volume={42},
  number={4},
  pages={139--1},
  year={2023}
}

@article{liu2023libero,
  title={Libero: Benchmarking knowledge transfer for lifelong robot learning},
  author={Liu, Bo and Zhu, Yifeng and Gao, Chongkai and Feng, Yihao and Liu, Qiang and Zhu, Yuke and Stone, Peter},
  journal={Advances in Neural Information Processing Systems},
  volume={36},
  pages={44776--44791},
  year={2023}
}

@article{nasiriany2024robocasa,
  title={Robocasa: Large-scale simulation of everyday tasks for generalist robots},
  author={Nasiriany, Soroush and Maddukuri, Abhiram and Zhang, Lance and Parikh, Adeet and Lo, Aaron and Joshi, Abhishek and Mandlekar, Ajay and Zhu, Yuke},
  journal={arXiv preprint arXiv:2406.02523},
  year={2024}
}

@article{bhat20253d,
  title={3d cavla: Leveraging depth and 3d context to generalize vision language action models for unseen tasks},
  author={Bhat, Vineet and Lan, Yu-Hsiang and Krishnamurthy, Prashanth and Karri, Ramesh and Khorrami, Farshad},
  journal={arXiv preprint arXiv:2505.05800},
  year={2025}
}

@article{luo2026being,
  title={Being-H0. 5: Scaling Human-Centric Robot Learning for Cross-Embodiment Generalization},
  author={Luo, Hao and Wang, Ye and Zhang, Wanpeng and Zheng, Sipeng and Xi, Ziheng and Xu, Chaoyi and Xu, Haiweng and Yuan, Haoqi and Zhang, Chi and Wang, Yiqing and others},
  journal={arXiv preprint arXiv:2601.12993},
  year={2026}
}

@article{yuan2026fast,
  title={Fast-WAM: Do World Action Models Need Test-time Future Imagination?},
  author={Yuan, Tianyuan and Dong, Zibin and Liu, Yicheng and Zhao, Hang},
  journal={arXiv preprint arXiv:2603.16666},
  year={2026}
}

@article{cen2025worldvla,
  title={Worldvla: Towards autoregressive action world model},
  author={Cen, Jun and Yu, Chaohui and Yuan, Hangjie and Jiang, Yuming and Huang, Siteng and Guo, Jiayan and Li, Xin and Song, Yibing and Luo, Hao and Wang, Fan and others},
  journal={arXiv preprint arXiv:2506.21539},
  year={2025}
}

@inproceedings{lu2024manigaussian,
  title={Manigaussian: Dynamic gaussian splatting for multi-task robotic manipulation},
  author={Lu, Guanxing and Zhang, Shiyi and Wang, Ziwei and Liu, Changliu and Lu, Jiwen and Tang, Yansong},
  booktitle={European Conference on Computer Vision},
  pages={349--366},
  year={2024},
  organization={Springer}
}

@article{yu2026forcevla,
  title={Forcevla: Enhancing vla models with a force-aware moe for contact-rich manipulation},
  author={Yu, Jiawen and Liu, Hairuo and Yu, Qiaojun and Ren, Jieji and Hao, Ce and Ding, Haitong and Huang, Guangyu and Huang, Guofan and Song, Yan and Cai, Panpan and others},
  journal={Advances in Neural Information Processing Systems},
  volume={38},
  pages={93409--93439},
  year={2026}
}

@inproceedings{wu2025momanipvla,
  title={Momanipvla: Transferring vision-language-action models for general mobile manipulation},
  author={Wu, Zhenyu and Zhou, Yuheng and Xu, Xiuwei and Wang, Ziwei and Yan, Haibin},
  booktitle={Proceedings of the Computer Vision and Pattern Recognition Conference},
  pages={1714--1723},
  year={2025}
}

@article{li2026bridgevla,
  title={Bridgevla: Input-output alignment for efficient 3d manipulation learning with vision-language models},
  author={Li, Peiyan and Chen, Yixiang and Wu, Hongtao and Ma, Xiao and Wu, Xiangnan and Huang, Yan and Wang, Liang and Kong, Tao and Tan, Tieniu},
  journal={Advances in Neural Information Processing Systems},
  volume={38},
  pages={63635--63673},
  year={2026}
}

@article{fan2026any3d,
  title={Any3D-VLA: Enhancing VLA Robustness via Diverse Point Clouds},
  author={Fan, Xianzhe and Deng, Shengliang and Wu, Xiaoyang and Lu, Yuxiang and Li, Zhuoling and Yan, Mi and Zhang, Yujia and Zhang, Zhizheng and Wang, He and Zhao, Hengshuang},
  journal={arXiv preprint arXiv:2602.00807},
  year={2026}
}

@inproceedings{teed2020raft,
  title={Raft: Recurrent all-pairs field transforms for optical flow},
  author={Teed, Zachary and Deng, Jia},
  booktitle={European conference on computer vision},
  pages={402--419},
  year={2020},
  organization={Springer}
}

@article{yang2024depth,
  title={Depth anything v2},
  author={Yang, Lihe and Kang, Bingyi and Huang, Zilong and Zhao, Zhen and Xu, Xiaogang and Feng, Jiashi and Zhao, Hengshuang},
  journal={Advances in Neural Information Processing Systems},
  volume={37},
  pages={21875--21911},
  year={2024}
}
